\documentclass[12pt]{colt2024} 

\title[Scale-free Adversarial Reinforcement Learning]{Scale-free Adversarial Reinforcement Learning }
\usepackage{times}

\coltauthor{%
 \Name{Mingyu Chen} \Email{mingyuc@bu.edu}\\
 \addr Boston University
 \AND
 \Name{Xuezhou Zhang} \Email{xuezhouz@bu.edu}\\
 \addr Boston University
}

\def\E{\mathbb{E}}
\def\P{\mathbb{P}}

\def\R{\mathbb{R}}
\def\p{\mathbf{p}}
\def\q{\mathbf{q}}

\usepackage{breqn}
\usepackage{multirow}
\usepackage{makecell}

\usepackage{tikz}
\newcommand*\circled[1]{\tikz[baseline=(char.base)]{
            \node[shape=circle,draw,inner sep=2pt] (char) {#1};}}
\newcommand{\mathbbm}[1]{\text{\usefont{U}{bbm}{m}{n}#1}} 
\SetKwInput{KwInput}{Input}
\SetKwInput{KwOutput}{Output}
\SetKwInput{KwInitialize}{Initialize}

\usepackage{setspace}

\begin{document}

\maketitle

\begin{abstract}%
This paper initiates the study of scale-free learning in Markov Decision Processes (MDPs), where the scale of rewards/losses is unknown to the learner. We design a generic algorithmic framework, \underline{S}cale \underline{C}lipping \underline{B}ound (\texttt{SCB}), and instantiate this framework in both the adversarial Multi-armed Bandit (MAB) setting and the adversarial MDP setting. Through this framework, we achieve the first minimax optimal expected regret bound and the first high-probability regret bound in scale-free adversarial MABs, resolving an open problem raised in \cite{hadiji2023adaptation}. On adversarial MDPs, our framework also give birth to the first scale-free RL algorithm with a $\tilde{\mathcal{O}}(\sqrt{T})$ high-probability regret guarantee.
\end{abstract}

\begin{keywords}%
  bandit, MDP, scale-free learning, high-probability regret.
\end{keywords}

\section{Introduction}

Reinforcement learning (RL) refers to the problem of an
agent interacting with an unknown environment with the
goal of improving its policy and minimizing cumulative loss through time.
The environment is commonly modeled as a Markov Decision
Process (MDP) with an unknown transition function.
In this paper we focus on the adversarial MDP setting, where the losses are allowed to be generated adversarially \citep{even2009online}.
Curiously, virtually \textit{all} prior works on RL assume that the rewards/losses are uniformly bounded, e.g. the mean reward for any state-action pair is within $[0,1]$.
This regularity condition is crucial in allowing existing algorithms to set their hyper-parameters such as learning rate properly to achieve low regrets.
In many real-world applications, however, such natural loss bound does not always exist. 
For instance, in quantitative trading, stock prices can vary significantly over time and across different stocks.
More importantly, the scale of such variance is often not known to the algorithm a priori. In such settings, most existing algorithms no longer work.

Motivated by the above limitations, in this paper, we initiate the study of scale-free RL algorithms in MDPs, i.e. algorithms that require no prior knowledge on the scale of the losses.
Scale-free algorithm have previously been studied in the online learning literature
\citep{freund1997decision, de2014follow, cesa2007improved, mayo2022scale, jacobsen2023unconstrained, cutkosky2019artificial}.
However, for decision-making under uncertainty, the only relevant studies are limited to Multi-armed Bandits (MAB) \citep{hadiji2023adaptation, putta2022scale, chen2023improved, huang2023banker}, which 
can be considered as a $1$-layer MDP with a single state.
Existing algorithms for scale-free MAB essentially adapts algorithms designed for online learning to the bandit feedback setting.
This leads to some fundamental problems.
For example, due to the limitations of regularizers, no existing scale-free adversarial MAB algorithms can achieve minimax optimality (i.e., optimal to logarithm terms).
Secondly, existing works only bound the regret incurred by the important weighted estimators.
As a result, they can only bound the expected regret and cannot be generalized to high probability regret.
Third, considering the existing algorithms' dependence on important weighted estimators, their results cannot be generalized to the setting of adversarial MDP with unknown transition function.

\begin{table}[t]
\begin{minipage}{\textwidth}
    \renewcommand{\arraystretch}{1.5}
    \resizebox{\textwidth}{!}{%
    \begin{tabular}{|c|c|c|c|c|}\hline
    Setting & Loss & Algorithm & Regret & Type  \\\hline
    \multirow{6}{*}{\makecell{Adversarial MABs}} & \multirow{2}{*}{Bound} & {\cite{audibert2009minimax}} & ${\Theta}(\ell_\infty\sqrt{nT})$ & Exp. \\\cline{3-5}
     & & \cite{neu2015explore} &  ${\Theta}(\ell_\infty\sqrt{nT\log(n/\delta)})$ & High prop. \\ \cline{2-5}
     &  \multirow{4}{*}{Unbound} & \cite{hadiji2023adaptation} &  ${\Theta}(\ell_\infty\sqrt{nT\log n})$ & Exp. \\\cline{3-5}
     & & \cite{chen2023improved} & ${\Theta}(\ell_\infty\sqrt{nT\log T})$ & Exp.  \\ \cline{3-5}
    & & \textbf{\texttt{SCB}} (\textbf{Theorem~\ref{theo:mmopt}}) & ${\Theta}(\ell_\infty\sqrt{nT})$ & Exp.  \\\cline{3-5}
     & & \textbf{\texttt{SCB-IX}} (\textbf{Theorem~\ref{theo:highprop}}) & ${\Theta}(\ell_\infty\sqrt{nT \log (n/\delta) })$ & High prop.  \\\cline{1-5}
     \multirow{2}{*}{\makecell{Adversarial MDPs}} & {Bound} & \cite{jin2019learning} & $\tilde{\mathcal{O}}(\sum_{h\in [H]} \ell_{\infty,h} S\sqrt{AT})$ & High prop.  \\\cline{2-5}
     & Unbound & \textbf{\texttt{SCB-RL}} (\textbf{Theorem~\ref{SCB_RL_1}}) & $\tilde{\mathcal{O}}(\sum_{h\in [H]} \ell_{\infty,h} S^{3/2}\sqrt{AT})$ & High prop.  \\\cline{1-5}
    \end{tabular}}
    \caption{An overview of the proposed algorithms/results and comparisons with related works.}
    \label{tab:related work}
\end{minipage}
\end{table}

In this paper, we propose the first scale-free algorithm for adversarial MDPs with unknown transition function.
We design a unified framework called \underline{S}cale \underline{C}lipping \underline{B}ound (\texttt{SCB}).
This framework can be applied to both MAB and MDP and significantly improves previous results across the board.
Our technical contributions can be summarized below, and an overview that compare our results with those in prior works can be found in Table \ref{tab:related work}.
\begin{enumerate}
    \item
    We propose \texttt{SCB}, a scale-free adversarial MAB algorithm that achieves \textbf{minimax optimal} expected regret bounds without the knowledge of the loss magnitude. 
    Our result eliminates the $\log(n)$ and $\log(T)$ factors in prior works, and matches the minimax lower-bound \cite{auer2002finite} upto constant factors.
    This result gives a positive answer to the open problem raised in \cite{hadiji2023adaptation}.
    \item 
    Based on the idea of \texttt{SCB}, we build \texttt{SCB-IX}, the {first} scale-free adversarial MAB algorithm that achieves a \textbf{high probability} regret bound.
    \item 
    Finally, we extend the above ideas to the setting of adversarial MDPs and present \texttt{SCB-RL}, the \textbf{first} scale-free algorithm that achieve $\tilde{\mathcal{O}}(\sqrt{T})$ high probability regret bound for adversarial MDP with unknown transition function, unbounded losses and bandit feedback.
\end{enumerate}

\section{Related Works}
\paragraph{Scale-free learning:}
Scale-free algorithms refer to the algorithms that do not need to know any upper or lower bounds on the loss functions. 
Scale-free regret bounds were first studied in the full information setting, such as experts problems \citep{freund1997decision, de2014follow, cesa2007improved} and online convex optimization \citep{mayo2022scale, jacobsen2023unconstrained, cutkosky2019artificial}.
For experts problems, the \texttt{AdaHedge} algorithm from \cite{de2014follow} achieves the first scale-free regret bound.
For online convex optimization, past algorithms can be categorized into two generic algorithmic frameworks: Mirror Descent (MD) and Follow The Regularizer Leader (FTRL).
The scale-free regret from the MD family is achieved by \texttt{AdaGrad} proposed by \cite{duchi2011adaptive}.
However, the regret bound of \cite{duchi2011adaptive} is only non-trivial when the Bregman divergence associated with the regularizer can be well bounded.
Later, the \cite{orabona2018scale} proposed the \texttt{AdaFTRL} algorithm which achieves the first scale-free regret bound in the FTRL family and generalizes \cite{duchi2011adaptive}'s results to cases where the Bregman divergence associated with the regularizer is unbounded.
For the adversarial MAB problem, \cite{hadiji2023adaptation} extends the method of \cite{duchi2011adaptive} and provides a scale-free regret bound of $\widetilde O\Big(\ell_{\infty}\sqrt{nT}\Big)$, which is optimal (up to log terms) in the worst case.
\cite{putta2022scale} design a bandit FTRL algorithm and presents scale-free bounds that adapt to the individual size of losses across time.
Unfortunately, the worst-case guarantee of \cite{putta2022scale} is $\widetilde O\Big(\ell_{\infty}n\sqrt{T}\Big)$, which scales linearly to the number of actions.
To close the gap, \cite{chen2023improved} proposes algorithms that achieves an adaptive regret better than \cite{putta2022scale}, as well as an optimal worst-case regret that matches with \cite{hadiji2023adaptation}.

Notice that all previous studies are unable to attain logarithmic optimality, e.g., the regret bound of \cite{putta2022scale, chen2023improved, huang2023banker} is $\Theta(\ell_\infty\sqrt{nT\log {T} })$, and the regret bound of \cite{hadiji2023adaptation} is $\Theta(\ell_\infty\sqrt{nT\log {n} })$.
This is due to some inherent limitations of their algorithms. 
To be more specific, \cite{hadiji2023adaptation} is an extension of \texttt{AdaHedge} \cite{de2014follow}, with a structure similar to \texttt{EXP3}, leading to an additional \(\sqrt{\log n}\) regret. 
On the other hand, the analysis of \cite{putta2022scale, chen2023improved, huang2023banker} is only applicable to algorithms with a log-barrier regularizer, which also results in an additional \(\sqrt{\log T}\) regret.
To achieve logarithmic optimality, a promising approach would be to use algorithms with Tsallis-INF regularizer \citep{audibert2009minimax}, which can achieve $\Theta(\ell_\infty\sqrt{nT})$ regret bound when $\ell_\infty$ is known.
However, when $\ell_\infty$ is unknown, it is unclear whether this bound can be achieved.
This has been posed as an open problem in \cite{hadiji2023adaptation} (Remark 8), which we answer in the positive.

\paragraph{High probability regrets:} 
High-probability regrets for adversarial MAB were first provided by \cite{auer2002nonstochastic} and explored in a more generic way by \cite{abernethy2009beating}.
The idea is to reduce the variance of importance weighted estimators by adding \textit{explicit exploration} on the action distribution.
Later, \cite{kocak2014efficient} and \cite{neu2015explore} improve the \textit{explicit exploration} method to \textit{implicit exploration}, and design algorithms for more complex models with potentially large action sets and side information.
Notably, all the above algorithms require carefully constructing \textit{biased} loss estimators.
In contrast, \cite{lee2020bias} develops algorithms based on \textit{unbiased} loss estimators, and enjoy \textit{data-dependent} high probability regret bounds, which could be much smaller than the bounds in the form of $\tilde{\mathcal{O}}(\sqrt{T})$ when the data is ``good''.
For the adversarial MDP problem, a rencent line of works develop algorithms with high-probability regret bounds \citep{jin2019learning, lee2020bias, luo2021policy, dai2022follow, jin2022near}.
Most of them are based on the idea of reducing an adversarial MDP problem to an adversarial MAB problem through \textit{occupancy measure} and then solve it using bandit algorithms, and achieve the same regret guarantee.
To the best of our knowledge, there are no studies on the high probability regret for either adversarial MAB or MDP with considering unbounded losses, a gap that we fill in this work.

{
\paragraph{Variance dependent regrets:} 
Variance dependent regrets have been studied in both adversarial MAB \citep{hazan2011better, bubeck2018sparsity, wei2018more, ito2021parameter} and MDP \citep{talebi2018variance, simchowitz2019non, zhang2023settling, zanette2019tighter}.
At first glance, it seems that a variance-dependent regret can automatically adapt to the scale of the losses, thereby directly implying scale-adaptive regret. 
However, in fact, there is a fundamental gap between the concept of scale-free and scale-adaptive. 
Firstly, scale-adaptive algorithms require a \textit{strict} assumption that the scale of losses can be bounded by a known constant $L$. This applies to all the above-mentioned work.
If the assumption is violated, their analyses will not hold. 
Secondly, the above results on variance-dependent regret all include a burn-in term that scales polynomial to $L$. 
This leads to the optimality of these results being guaranteed only in a ``large-sample'' regime. 
As $L$ goes towards infinity, the burn-in term eventually dominates.
}

\section{Adversarial Multi-armed Bandit}
Let us start our discussion with adversarial MAB.
The scale-free MAB problem proceeds in rounds between a player and an adversary. 
In each round $t=1,\dots, T$, the player selects one of the $n$ available actions $k_t\in [n]$, while the adversary at the same time picks a loss vector $\ell_t \in \R^n$ with $\ell_{t, k}$ being the loss for action $k$.
We assume the adversary is \textit{adaptive}: the adversary can choose $\ell_t$ base on the player's previous actions in an arbitrary way.
At the end of round $t$, the learner observes the loss of the chosen action $\ell_{t, k_t}$ and nothing else. We measure the scale of the losses by $\ell_\infty = \max_{t\in [T], k\in [n]}|\ell_{t,k}|$. 
We measure the performance of the learner in terms of its \textit{regret}:
\begin{align*}
    \mathcal{R}(T) = \sum_{t=1}^T \ell_{t, k_t} - \min_{k\in [n]} \sum_{t=1}^T \ell_{t, k}.
\end{align*}

\subsection{Minimax Optimal Expected Regret}
In this subsection we focus on bounding the \textit{expected regret}, i.e., $\E[\mathcal{R}(T)]$.
Compared to existing works, there are several important advantages of our approach: 
1). Our algorithm archives the first minimax optimal expected regret $\Theta(\ell_\infty\sqrt{nT})$, which significantly improves upon the $\Theta(\ell_\infty\sqrt{nT\log {T} })$ results in \cite{putta2022scale, chen2023improved, huang2023banker} and $\Theta(\ell_\infty\sqrt{nT\log {n} })$ in \cite{hadiji2023adaptation}, and matches the lower bound $\Omega(\ell_\infty\sqrt{nT})$ proposed in \cite{auer2002nonstochastic}. 
2). Our algorithm is \textit{strongly scale-free} \citep{orabona2018scale}, that is, with the same parameters, the sequence of action distributions of the algorithm does not change if the sequence of loss is multiplied by a positive constant. 
Such property is previously implemented only in \cite{hadiji2023adaptation}. 

Our design is illustrated in Algorithm~\ref{ALG1}.
The algorithm follows a standard Follow-the-regularized-Leader (FTRL) framework.
At the beginning of round $t$, the algorithm computes an action distribution $\p_t\in \Delta_n$ such that
\begin{align}
\label{FTRL::update_rule}
    \p_t =\arg\min_{\p\in \Delta_n} \left( \sum_{s=1}^{t-1}\langle \hat \ell_s, \p \rangle + \frac{1}{\eta_{t}}\Psi(\p) \right),
\end{align}
where $\hat \ell_s$ is an estimator of $\ell_s$ and $\Psi$ is the regularizer.
Then, the algorithm derives $\q_t$ by mixing $\p_t$ with a uniform distribution, samples and plays action $k_t\sim \q_t$, and obtains loss $\ell_{t,k_t}$.
The key of our design lies in the construction of the loss estimator.
In round $t$, the algorithm holds a ``scale clipping bound'' (i.e., clipping threshold) $C_t$, which is twice the largest scale among the previously observed losses.
After receiving the loss $\ell_{t,k_t}$, the algorithm clips the loss within the interval $[-C_t, C_t]$, incorporates an offset $C_t$ to make the loss non-negative, and construct the importance-weighted loss estimator, i.e., $\hat\ell_{t,k} = (\max(-C_t, \min(C_t, \ell_{t,k}))+C_t)\mathbbm{1}\{k=k_t\}/q_{t,k}$.
Specifically, notice that $C_t$ is independent to $\p_t$, thereby $\hat \ell_{t,k}$ is an unbiased estimator of $\max(-C_t, \min(C_t, \ell_{t,k}))+C_t$.
At the end of round $t$, the algorithm updates parameters $C_{t+1}, \eta_{t+1}, \beta_{t+1}$, and then move to the next round.
\begin{algorithm2e}[!t]
\label{ALG1}
\begin{spacing}{1}
    \DontPrintSemicolon  
  \KwInput{$1/2$-Tsallis Entropy $\Psi$, $\eta_1=\infty$, $\beta_1 = n/(2n+\sqrt{n})$, $C_1=0$}
 \For{$t=1,\dots,T$}
 {
    Compute the action distribution $\p_t =\arg\min_{\p\in \Delta_n} \left( \sum_{s=1}^{t-1}\langle \hat \ell_s, \p \rangle + \frac{1}{\eta_{t}}\Psi(\p) \right)$\;
    Add extra exploration $\q_t = (1-\beta_t)\p_t + \beta_t \frac{\mathbf{1}_n}{n}$\;
    Sample and play action $k_t\sim \q_t$. Receive loss $\ell_{t, k_t}$\;
    Clip received loss by $[-C_t, C_t]$: $\ell_{t,k}^c = \max(-C_t, \min(C_t, \ell_{t,k}))$\;
    Construct estimator $\hat\ell_t$ such that $\hat \ell_{t,k} =   \frac{\ell_{t,k}^c+C_t}{q_{t,k}}\mathbbm{1}\{k=k_t\},\ \forall k\in [n]$\;
    If $|\ell_{t,k_t}|>C_t$, set $C_{t+1} = 2 |\ell_{t,k_t}|$, otherwise $C_{t+1} = C_t$\;
    Update learning rate $\eta_{t+1} =  \frac{1}{2C_{t+1}\sqrt{t+1}}$. Update exploration rate $\beta_{t+1} 
 =\frac{n}{2n+\sqrt{{n(t+1)}}}$\;
 }
\caption{SCB: Scale Clipping Bound}
\end{spacing}
\end{algorithm2e}
The regret guarantee of Algorithm~\ref{ALG1} can be summarized below.
\begin{theorem}
    \label{theo:mmopt}
    Algorithm~\ref{ALG1} achieves 
    \begin{align*}
    \E[\mathcal{R}(T)]\le \Theta\left(\ell_\infty(n+\sqrt{nT})\right).
\end{align*}
\end{theorem}
\begin{remark}
\label{remark:1}
    We emphasize that \texttt{SCB} is \textbf{strongly scale-free}.
    When the sequence of losses is multiplied by a positive constant, the clipping threshold will also be rescaled accordingly, resulting in the distributions of actions not changing. This property is also inherited by the derived algorithms \texttt{SCB-IX} and \texttt{SCB-RL}.
    More details about this property are provided in Appendix~\ref{apendix:remark:1}.
\end{remark}

\noindent\textbf{Proof Sketch}:
Denoted by $\ell_{t,k}^c = \max(-C_t, \min(C_t, \ell_{t,k}))+C_t)$, we start with the following regret decomposition. 
\begin{align*}
    \E\left[\mathcal{R}(T)\right] &= \E\left[\sum_{t=1}^T \langle \ell_t, \q_t-\p^* \rangle\right]\\
    &= \E\left[\sum_{t=1}^T \langle \ell_t^c, \p_t-\p^* \rangle\right]+ \E\left[\sum_{t=1}^T \langle \ell_t, \q_t-\p_t \rangle\right]+\E\left[\sum_{t=1}^T \langle \ell_t-\ell_t^c, \p_t-\p^* \rangle\right]\\
    &= \E\left[\sum_{t=1}^T \langle \ell_t^c+ C_t\mathbf{1}_n, \p_t-\p^* \rangle\right] + \E\left[\sum_{t=1}^T \langle \ell_t, \q_t-\p_t \rangle\right]+\E\left[\sum_{t=1}^T \langle \ell_t-\ell_t^c, \p_t-\p^* \rangle\right]\\
    & = 
    \underbrace{\E\left[\sum_{t=1}^T \langle \hat \ell_t, \p_t-\p^* \rangle\right]}_{\circled{1}} 
    + 
    \underbrace{\E\left[\sum_{t=1}^T \langle \ell_t, \q_t-\p_t \rangle\right]}_{\circled{2}}
    +
    \underbrace{\E\left[\sum_{t=1}^T \langle \ell_t-\ell_t^c, \p_t-\p^* \rangle\right]}_{\circled{3}}
\end{align*}
Here, $\p^*$ denotes the optimal comparator, which can be dependent on the algorithm's actions $k_1,\dots,k_T$.
The third equality is due to $\langle \mathbf{1}_n, \q_t-\p^* \rangle=0$, and the last equality is because $\hat\ell_t$ is an unbiased estimator of $\ell_t^c+ C_t\mathbf{1}_n$.
Here, term \circled{1} is the regret of the corresponding FTRL algorithm;
term \circled{2} corresponds to the error incurred by mixing with uniform distribution; 
term \circled{3} measures the error of the clipping.

\noindent\textbf{Bounding \circled{1}}:
We first bound the FTRL regret.
The proof is founded on an observation that $0\le \ell_{t,k_t}^c+C_t\le 2 C_t$ for every $t\in [T]$, where $C_t$ is a value known to the algorithm at the beginning of round $t$.
In this case, we can tune the learning rate to fit the scale of the loss before observing it, thereby reducing the analysis to the bounded case.
The main result is as follows.
The detailed proof is delayed to Appendix~\ref{appendix:lem:minimax_1}.
\begin{lemma}
\label{lem:minimax_1}
Algorithm~\ref{ALG1} ensures 
    \begin{align*}
        \E\left[\sum_{t=1}^T \langle \hat \ell_t, \p_t-\p^\star \rangle\right]\le \Theta\left(\ell_\infty \sqrt{nT}\right),
    \end{align*}
    where $\ell_\infty = \max_{t\in [T], k\in [n]}|\ell_{t,k}|$.
\end{lemma}

\noindent\textbf{Bounding \circled{2}}: The proof is trivial since
\begin{align*}
     \E\left[\sum_{t=1}^T \langle \ell_{t}, \q_t-\p_t \rangle\right]= \E\left[\sum_{t=1}^T \beta_t\langle  \ell_{t}, \frac{\mathbf{1}_n}{n}-\p_t \rangle \right]
     \le 2\ell_\infty \sum_{t=1}^T \frac{n}{2n+\sqrt{nt}}  \le 4\ell_\infty\sqrt{nT} .
\end{align*}

\noindent\textbf{Bounding \circled{3}}:
Bounding the clipping error is the key to the entire proof.
Define $K:=\arg\min_{j\in \mathbb{N}}\left\{  \ell_\infty \le 2^j \right\}$.
Define $\ell_t^i\in \R^n$ such that $\ell_{t,k}^i = \ell_{t,k}\mathbbm{1}\{2^{i-1}<|\ell_{t,k}|\le 2^i\}$ for $k\in [n]$.
Notice that $\ell_t = \sum_{i=-\infty}^K \ell_t^i$.
In this case, there is 
\begin{align*}
    \E[\sum_{t=1}^T \langle \ell_t - \ell_t^c, \p_t-\p^\star  \rangle]\le 2\E\left[\sum_{t=1}^T  \|\ell_t - \ell_t^c\|_\infty\right]\le 2\E\left[\sum_{i=-\infty}^K\E\left[\sum_{t=1}^{T} \|\ell_t^i - {\ell^c_t}^i\|_\infty\right]\right].
\end{align*}
We focus on the inner terms.
We first note
\begin{align*}
    \E\left[\sum_{t=1}^{T} \|\ell_t^i - {\ell_t^c}^i\|_\infty\right]\le \E\left[2^i \sum_{t=1}^{T} \mathbbm{1}\{\ell_t^i\not=\mathbf{0}_n\} \mathbbm{1}\{C_t< 2^i\} \right] 
\end{align*}
since the clipping threshold is non-decreasing and all non-zero entries in $\ell_t^i$ are within $[2^{i-1}, 2^i]$.
Now it suffices to bound $\E[ \sum_{t=1}^{T} \mathbbm{1}\{\ell_t^i\not=\mathbf{0}_n\} \mathbbm{1}\{C_t< 2^i\}]$.
To this end, an important observation is that for every integer $m\ge 1$, there is
\begin{align}
\label{eq:clipErr}
    \P\left\{ \sum_{t=1}^{T} \mathbbm{1}\{\ell_t^i\not=\mathbf{0}_n\} \mathbbm{1}\{C_t< 2^i\} \ge m \right\}\le \left(1-\frac{\beta_T}{n}\right)^{m-1}.
\end{align}
This is because $\sum_{t=1}^{T} \mathbbm{1}\{\ell_t^i\not=\mathbf{0}_n\} \mathbbm{1}\{C_t< 2^i\} \ge m$ implies that the algorithm does not play the non-zero entries of the first $m-1$ non-zeros losses $\ell_t^i$.
Otherwise, in the $m$-th round where $\ell_{t}^i$ is non-zero, the clipping threshold should be no less than $2^i$, which implies that no clipping can happen.
Since each action has probability at least $\beta_t/n\ge \beta_T /n$ to be played every round, \eqref{eq:clipErr}  immediately follows.
Thus, there is 
\begin{align*}
    \E\left[ \sum_{t=1}^{T} \mathbbm{1}\{\ell_t^i\not=\mathbf{0}_n\} \mathbbm{1}\{C_t< 2^i\} \right] &= \sum_{m=1}^\infty  \P\left\{ \sum_{t=1}^{T} \mathbbm{1}\{\ell_t^i\not=\mathbf{0}_n\} \mathbbm{1}\{C_t< 2^i\} \ge m \right\}\\
    &\le \frac{n}{\beta_T} = 2n+\sqrt{nT}.
\end{align*} 
Now we take the sum of all $i\le K$.
\begin{align*}
    \E\left[\sum_{t=1}^T \langle \ell_t - \ell_t', \p_t-\p^\star \rangle\right]&\le 2\E\left[\sum_{i=-\infty}^K\E\left[\sum_{t=1}^{T} \|\ell_t^i - {\ell_t^c}^i\|_\infty\right]\right]\\
    &\le 2\E\left[\sum_{i=-\infty}^K 2^i (2n+\sqrt{nT})\right]\\
    &\le 2^{K+2}  (2n+\sqrt{nT})\le 8 \ell_\infty (2n+\sqrt{nT}).
\end{align*}
The last inequality is due to $2^{K-1}\le \ell_\infty$.
Combining \circled{1},\circled{2} and \circled{3}, we have 
\begin{align*}
    \E\left[\mathcal{R}(T)\right]\le \Theta\left(\ell_\infty(n+\sqrt{nT})\right),
\end{align*}
which is optimal upto constant factors.

\subsection{High probability regret}
Next, we study the more challenging problem of high-probability regret.
The goal is to design algorithms for which $\mathcal{R}(T)$ can be bounded with high probability.
We propose the first scale-free adversarial MAB algorithm with a high-probability regret guarantee.

The algorithm \texttt{SCB-IX} is provided in Algorithm~\ref{Alg2}.
Conceptually, the algorithm is a variant of \texttt{EXP3-IX} in \cite{neu2015explore} combined with the clipping idea in Algorithm~\ref{ALG1}. 
By Hoeffding's inequality, it suffices to focus on bounding $\sum_{t=1}^T \langle \ell_t, \q_t-\p^\star \rangle$.
Similar to the proof of Theorem~\ref{theo:mmopt}, we can decompose the regret into $\sum_{t=1}^T \langle \ell_t^c+C_t\mathbf{1}_n, \q_t-\p^\star  \rangle$ and $\sum_{t=1}^T \langle \ell_t - \ell_t^c, \q_t-\p^\star  \rangle$.
For the first term, due to $0\le \ell_{t,k}^c+C_t\le 2C_t $, where $C_t$ is known at the beginning of round $t$, it suffices to show that the regret can be well bounded with high probability based on the proof of \texttt{EXP3-IX}.
For the second term, as shown in inequality (\ref{eq:clipErr}), we have $ \sum_{t=1}^{T} \mathbbm{1}\{\ell_t^i\not=\mathbf{0}_n\} \mathbbm{1}\{C_t< 2^i\} \ge \log(1/\delta)n/\beta_T$ with probability at least $1-\delta$, which immediately imply a high probability bound for the clipping error.
Our results can be summarized in the following theorem.
\begin{theorem}
\label{theo:highprop}
    With probability at least $1-\delta$, Algorithm~\ref{Alg2} ensures 
    \begin{align*}
        \mathcal{R}(T)\le \Theta\left( \ell_\infty   \sqrt{\frac{n^2+nT}{\log n}}\log(1/\delta)  + \ell_\infty\sqrt{ nT \log (n)}\right),
    \end{align*}
\end{theorem}
Due to the space limit, the detailed proof is delayed to Appendix~\ref{appendix: high_prop}.
Specifically, when $T\ge n$, the regret reduces to $\Theta( \ell_\infty   \sqrt{{nT}/{\log n}}\log(1/\delta)  + \ell_\infty\sqrt{ nT \log (n)})$\footnote{Note that the bound scales linearly with $\log(1/\delta)$ for all levels $\delta$. The dependence can be improved to $\sqrt{\log(1/\delta)}$ if the algorithm use $\delta$ to tune its parameter \citep{neu2015explore}.
This is the way to derive the results presented in Table~\ref{tab:related work}.}.
This matches the results in \cite{neu2015explore} for the bounded loss setting.

\begin{algorithm2e}[!t]
\label{Alg2}
\begin{spacing}{1}
    \DontPrintSemicolon  
 \KwInput{Shannon Entropy $\Psi$, $\eta_1=\infty$, $\gamma_1=0$,   $\beta_1 = {\sqrt{n \log (n)/ (n \log (n)+1)}}$, $C_1=0$}
 \For{$t=1,\dots,T$}
 {
    Compute the action distribution $\p_t =\arg\min_{\p\in \Delta_n} \left( \sum_{s=1}^{t-1}\langle \hat \ell_s, \p \rangle + \frac{1}{\eta_{t}}\Psi(\p) \right)$\;
    Add extra exploration $\q_t = (1-\beta_t)\p_t + \beta_t \frac{\mathbf{1}_n}{n}$\;
    Sample and play action $k_t\sim \q_t$. Receive loss $\ell_{t, k_t}$\;
    Clip received loss by $[-C_t, C_t]$: $\ell_{t,k}^c = \max(-C_t, \min(C_t, \ell_{t,k}))$\;
    Construct estimator $\hat\ell_t$ such that $\hat \ell_{t,k} =   \frac{\ell_{t,k}^c+C_t}{q_{t,k}+\gamma_t}\mathbbm{1}\{k=k_t\},\ \forall k\in [n]$\;
    If $|\ell_{t,k_t}|>C_t$, set $C_{t+1} = 2|\ell_{t,k_t}|$, otherwise $C_{t+1} = C_t$\;
    Update learning rate $\eta_{t+1} =  \frac{1}{C_{t+1}} \sqrt{\frac{\log n}{ n(t+1)}}$. Update exploration rate $\beta_{t+1} = \sqrt{\frac{n\log n }{n\log n+ t+1}}$, $\gamma_{t+1} = \eta_{t+1}C_{t+1}/2$\;
 }
\caption{SCB-IX: Scale Clipping Bound with Implicit Exploration}
\end{spacing}
\end{algorithm2e}

\section{Adversarial Markov Decision Process} 
With our preparation in the MAB setting, we now turn our attention to adversarial MDPs.
We consider the episodic MDP setting with finite horizon, unknown transition matrix, bandit feedback, and adversarial losses, same as the setting in \citet{jin2019learning}.
However, unlike \citet{jin2019learning}, where the losses are assumed to be in $[0,1]$, we allow the losses to be \textit{unbounded}.
To the best of our knowledge, this is the first study of RL with unbounded losses.

An adversarial MDP is defined by a tuple $(S,A,P,\{\ell_t\}_{t=1}^T)$.
$S$ is the finite state space and $A$ is the finite action space.
$P: S \times A\times S\to [0,1]$ is an unknown transition function where $P(s'|s,a)$ is the probability of reaching state $s'$ after taking action $a$ at state $s$.
$\ell_t: S\times A\to \R$ is a loss function determined by the adversary, which can depend on the player's actions before $t$.
Learning proceeds in $T$ episodes. 
In each episode $t$, the learner starts from state
$x_1$ and deploys a stochastic policy $\pi_t\in \Pi: S\times A\to [0,1]$ with $\pi_t(a|s)$ being the probability of taking action $a$ at state $s$.
The learner observes a state-action-loss trajectory $(s_1, a_1, \ell_t(s_1, a_1), \dots, s_{H}, a_{H}, \ell_t(s_{H}, a_{H}))$ before reaching the ending state $s_{H+1}$.
With a slight abuse of notation, we assume $\ell_t(\pi) = \E[\sum_{h\in [H]} \ell_t(s_h, a_h)|P, \pi]$.
The performance is measured by the regret, which is defined by 
\begin{align*}
    \mathcal{R}(T) = \sum_{t=1}^T \ell_t(\pi_t) - \min_{\pi\in \Pi} \sum_{t=1}^T \ell_t(\pi).
\end{align*}
Without loss of generality, we consider a layered structure MDP: the state space is partitioned into $H+2$ horizons $S_0,\dots, S_{H+1}$ such that $S=\cup_{h=1}^{H} S_h$, $\emptyset = S_i\cap S_j$ for every $i\not=j$, $S_0=\{s_0\}$ and $S_{H+1}=\{s_{H+1}\}$.
We further assume that the number of states in each horizon is the same, i.e., $S_h = S/H$ for all $h=[H]$.
Given the structure, with the help of ``occupancy meansure'' concept, this problem can be  restructured in a way that makes it highly similar to adversarial MAB: denoted the probability that policy $\pi$ visits the state-action pair $(s,a)$ with transition function $P$ by $q^{P, \pi}(s,a)$, the loss can be expressed as $\ell_t(\pi) = \sum_{s\in [S]}\sum_{a\in [A]} q^{P, \pi}(s,a)\ell_t(s,a) = \langle q^{P, \pi}, \ell_t  \rangle$.

While we have formulated the loss function in a form similar to the ones in adversarial MAB with the help of occupancy measure, a significant distinction still exists, which also constitutes the main challenge: for adversarial MDP, there is no explicit ``exploration policy'' guaranteeing that every state can be visited.
In particular, some states may be hardly accessible by any policy.
In such cases, directly implementing the proposed scale-free MAB algorithms would result in unbounded clipping errors, as the algorithm is unable to detect the scale changes in states that are not accessible.
In order to design scale-free algorithms for adversarial MDP, two critical questions need to be addressed: 1). How to find a good exploration policy for every state within $o(T)$ episodes? 2). How to handle the states that are hardly accessible for all policies?

To address these two questions, we design an exploration algorithm \texttt{RF-ELP}, as shown in Algorithm~\ref{RF-ELP}.
Conceptually, for each state $s$ that is accessible by some policy with a probability exceeding $\mathcal{\tilde O}(H\sqrt{SA/T})$, i.e., $\max_{\pi\in \Pi} q^{P, \pi}(s)\ge \mathcal{\tilde O}(H\sqrt{SA/T})$, \texttt{RF-ELP} is capable of producing a policy $\pi^{s,N}$ that successfully visit the state $s$ at least once every $\mathcal{O}(\sqrt{ST/A \max_{\pi\in \Pi} q^{P, \pi}(s) })$ episodes.
Additionally, for those states that are inaccessible by \texttt{RF-ELP}, we demonstrate that the maximum regret incurred by such a state can be bounded by $\mathcal{\tilde O}(\sum_{h\in [H]} \ell_{\infty, h}\sqrt{SAT})$.
More details are provided in the next paragraph and the appendix.
\texttt{RF-ELP} allows us to effectively reduces the problem of scale-free adversarial MDP to that of scale-free adversarial bandits.
Building upon \texttt{RF-ELP} and \texttt{UOB-REPS} in \cite{jin2019learning}, we develop the main algorithm \texttt{SCB-RL} and subalgorithm \texttt{UOB-REPS-EX}. 
The pseudocode of \texttt{SCB-RL} is presented in Algorithm~\ref{SCB-RL}, and the pseudocode of \texttt{UOB-REPS-EX} is delayed to Appendix~\ref{appendix:UOB-REPS-EX}.

Specifically, \texttt{SCB-RL} starts by calling \texttt{RF-ELP} for $\xi ST$ episodes and obtains an exploration policy for each of the states.
Then, in every episode $t$, it calls the subalgorithm \texttt{UOB-REPS-EX} to learn policy $\pi_t$, plays $\pi_t$ and receives a trajectory, clips and adds offset on the loss, updates the clipping threshold, and sends the information back to \texttt{UOB-REPS-EX}.
The subroutine \texttt{UOB-REPS-EX} is a variant of \texttt{UOB-REPS}, incorporating multiple designs for dealing with the unbounded losses, such as mixing with exploration policies and tuning the learning rate with the clipping threshold.
More details about \texttt{UOB-REPS-EX} are provided in Appendix~\ref{appendix:UOB-REPS-EX}.
The main theorem for \texttt{SCB-RL} is below.

\begin{theorem}
\label{SCB_RL_1}
    With probability at least $1-\delta$, \texttt{SCB-RL} guarantees
    \begin{align*}
        \mathcal{R}(T) \le \mathcal{\tilde O} 
 \left(\sum_{h\in [H]} \ell_{\infty, h} S^{3/2} \sqrt{AT}\right).
    \end{align*}
    where we denote $\ell_{\infty, h}=\max_{t\in [T],s\in [S_h], a\in [A]}\ell_t(s,a)$.
\end{theorem}

\begin{remark}
\label{remark:3}
    Compared to the best existing results with bounded losses $\mathcal{\tilde O} (\sum_{h\in [H]} \ell_{\infty, h} S \sqrt{AT})$, Theorem~\ref{SCB_RL_1} achieves the same optimality in terms $A, T$ but worse by a factor of \(\sqrt{S}\).
    Nevertheless, this result is quite surprising, considering that we spend additional episodes to learn the exploration policy for every state in \texttt{RF-ELP}.
    Furthermore, if all states are visitable, for sufficiently large $T$,
    we can further reduce the regret of \texttt{SCB-RL} to $\mathcal{\tilde O} (\sum_{h\in [H]} \ell_{\infty, h} S \sqrt{AT})$ by designing an early stopping strategy on \texttt{RF-ELP}, matching the best known regret in the bounded loss setting.
    We delay the details of this extension to Appendix~\ref{appendix:remark:3}.
\end{remark}

\noindent\textbf{Proof Sketch}: 
We start with the exploration algorithm \texttt{RF-ELP}.
As illustrated in Algorithm~\ref{RF-ELP}, the goal of \texttt{RF-ELP} is to find a set of policies each capable of visiting a particular state $s$ within $N$ episodes.
\texttt{RF-ELP} is essentially a reward-free exploration algorithm with a similar structure to that in \cite{jin2020reward}, while we replace the RL algorithm used for exploration from \texttt{EULER} \citep{zanette2019tighter} to \texttt{MVP} \citep{zhang2023settling}.
Specifically, \texttt{RF-ELP} starts by defining the reward $r^s$ as $r^s(s',a')=1$ if and only if $s'=s$, and then run \texttt{MVP} for $N$ episodes and get policy $\pi^{s, N}$.
Following this, \texttt{RF-ELP} resets the action distribution for state $s$ to ensure accessibility for every action.
We first present the theoretical guarantee of \texttt{MVP} as follows.

\begin{algorithm2e}[!t]
\label{RF-ELP}
\DontPrintSemicolon  
  \KwInput{State $s$; Exploration episodes number $N$}
  \KwOutput{Policy $\pi\in \Pi$}
  Initialize reward: $r^s(s', a') \leftarrow \mathbbm{1}\{s'=s\}$ for all $(s',a')\in [S]\times A$\;
  Run \texttt{MVP} \citep{zhang2023settling} $N$ episodes, get policies: $\{\pi^{s}_{1},\dots,\pi^s_N\} \leftarrow \texttt{MVP}(r^s, N)$, set $\pi^{s, N}\gets \text{Uniform}(\pi^{s}_{1},\dots,\pi^s_N)$\;
  Set policy $\pi^{s, N}(\cdot|s) \leftarrow \text{Uniform}(A)$\;
  Return $\pi^{s, N}$\;
\caption{Reward free exploration in RL (\texttt{RF-ELP})}
\end{algorithm2e} 

\begin{algorithm2e}[t]
\begin{spacing}{1}
\label{SCB-RL}
\DontPrintSemicolon  
  \KwInput{state space $S$, action space $A$, episode number $T$, state exploration parameter $\xi$}
    \KwInitialize{Clipping threshold $C_{1,h}=0$ for $h\in [H]$}
  \For{$s\in [S]$}
  {
    Run \texttt{RF-ELP} and update exploration policy: $\pi^{s} \leftarrow \texttt{RL-ELP}(s, \xi T )$\;
  }
  Send extra exploration policies $\{\pi^s\}_{s\in [S]}$ to \texttt{UOB-REPS-EX}\;
  \For{$t =\xi S T +1 \ \textbf{to} \ T$}
  {
  Receive policy $\pi_{t}\gets \texttt{UOB-REPS-EX}$\;
    Execute policy $\pi_t$ for $H$ horizons and obtain trajectory $\{s_h, a_h, \ell_t(s_h, a_h)\}_{h\in [H]}$\;
    Clip received loss by $[-C_{t,h}, C_{t,h}]$: 
    $\ell_t^c(s_h,a_h) = \max\left(-C_{t,h}, \min(C_{t,h},\ell_t(s_h,a_h))\right),\ \forall h\in [H]$\; 
        Send trajectory $\{s_h, a_h,\ell_t^c(s_h,a_h)+C_{t,h}\}_{h\in [H]}$ and clipping threshold $\{C_{t,h}\}_{h\in [H]}$ to \texttt{UOB-REPS-EX}.\;
    If $|\ell_t(s_h, a_h)|>C_{t,h}$, set $C_{t+1,h}=2|\ell_t(s_h, a_h)|$, otherwise $C_{t+1,h}=C_{t,h},\ \forall h\in [H]$\;
  }
\caption{\texttt{SCB-RL}: Scale Clipping Bound for RL}
\end{spacing}
\end{algorithm2e} 

\begin{lemma}[Theorem $3$ of \cite{zhang2023settling}]    \footnote{Notice that in this work $S$ represents the collection of states in all horizons,  corresponding to $SH$ in \cite{zhang2023settling}.}
\label{MVPg}
    For any $N\ge 1$ and $s\in [S]$, with probability at least $1-\delta$, \texttt{MVP} obeys 
    \begin{align*}
          \max_{\pi\in \Pi} \E\left[  \sum_{h\in [H]}  r^s(s_h, a_h) |P, \pi \right]-\E\left[  \sum_{h\in [H]}  r^s(s_h, a_h) |P, \pi^{s,N} \right]\le \mathcal{\tilde O}\left( \sqrt{\frac{SA \textup{Var}^s }{N}}+ \frac{SAH}{N}\right)
    \end{align*}
    where
    $
        \textup{Var}^s = \max_{\pi\in \Pi} \textup{Var}\left[ \sum_{h\in [H]} r^s(s_h,a_h) |P, \pi\right].
    $
\end{lemma}
Based on Lemma~\ref{MVPg}, we can derive the theoretical guarantee of \texttt{RF-ELP}. 
The key observation is that $\textup{Var}^s$ can be bounded by $\max_{\pi\in \Pi} \E[  \sum_{h\in [H]}  r^s(s_h, a_h) |P, \pi ]$ due to $ \sum_{h\in [H]}   r^s(s_h, a_h)\le 1$.
By the setting of $r^s$, it suffices to note that $ \E[  \sum_{h\in [H]}   r^s(s_h, a_h) |P, \pi ] = q^{P, \pi}(s)$ for every $\pi\in \Pi$.
In this case, when $\max_{\pi\in \Pi} q^{P,\pi}(s)\ge  \mathcal{\tilde O}({SAH}/{N})$, $\max_{\pi\in \Pi} q^{P, \pi}(s)$ and $q^{P, \pi^{s,N}}(s)$ should be on the same order.
We summarize the result in the following lemma.
\begin{lemma}[\texttt{RF-ELP} guarantee]
\label{lem:rfelp}
    For any $N\ge 1$ and $s\in [S]$, with probability at least $1-\delta$, if 
    $\max_{\pi\in \Pi} q^{P,\pi}(s)> 9C^2\left(\frac{SAH}{N}\right)$, then we have $q^{P,\pi^{s,N}}(s)\ge \frac{1}{2} \max_{\pi\in \Pi} q^{P,\pi}(s)$, where $C = \mathcal{O}(\log^3(T)\log(SA)\log(  \frac{ SAHT}{\delta}))$ is a poly-log factor w.r.t. $S,A, H, N$ and $1/\delta$.
\end{lemma}

The proof of Lemma~\ref{lem:rfelp} is proposed in Appendix~\ref{appendix:lem:rfelp}.
Now we begin to prove Theorem~\ref{SCB_RL_1}.
As illustrated in Algorithm~\ref{SCB-RL}, \texttt{SCB-RL} calls the exploration algorithm \texttt{RF-ELP} in the first $\xi S T$ episodes. 
For simplicity of the proof, we let $t$ start from $-\xi ST+1$ instead of $1$.
Denoted by $q_t = q^{P,\pi_t}$, as in \cite{jin2019learning}, the total regret can be written as $\sum_{t=-\xi ST+1}^T \langle \ell_t, q_t-q^*\rangle$.
We first decompose the regret into
\begin{align*}
    \sum_{t=-\xi ST+1}^T \langle \ell_t, q_t-q^*\rangle &={\sum_{t=-\xi ST+1}^{0} \langle \ell_t, q_t-q^*\rangle}+ {\sum_{t=1}^T \langle \ell_t^c, q_t-q^*\rangle} +{\sum_{t=1}^T \langle \ell_t-\ell_t^c, q_t-q^*\rangle}\\
    &\le \sum_{h\in [H]} \ell_{\infty, h} \xi ST+\underbrace{\sum_{t=1}^T \langle  \ell_t^+, q_t-q^*\rangle}_{\circled{1}} +\underbrace{\sum_{t=1}^T \langle \ell_t-\ell_t^c, q_t-q^*\rangle}_{\circled{2}},
\end{align*}
where $\ell_t^c(s,a)\in \R^{SA}$ and $\ell_t^+(s,a)\in \R^{SA}_+$ satisfy $\forall (s,a)\in [S]\times [A]$ \footnote{$h(s)$ is the index of the layer to which $s$ belongs.},
\begin{align*}
    \ell_t^c(s,a) &=  \max\left(-C_{t, h(s)}, \min(C_{t, h(s)},\ell_t(s,a))\right),\\
    \ell_t^+(s,a) &=\ell_t^c(s,a)+C_{t,h(s)}.
\end{align*}

\noindent\textbf{Bounding \circled{1}}:
The regret of \circled{1} is incurred by \texttt{UOB-REPS-EX}.
Similarly to \texttt{SCB-IX}, our algorithm differs to \texttt{UOB-REPS} in two key aspects: the mixing of explicit exploration and the presence of loss within the range of $[0, 2C_t]$ rather than $[0,1]$.
The result is stated below and the detailed proof is delayed to Appendix~\ref{appendix:lemma:9}.
\begin{lemma}
\label{lemma:9}
With probability at least $1-\delta$, there is
·    \begin{align*}
    \sum_{t=1}^T \langle \ell_t^+, q_t-q^*\rangle \le \mathcal{O}\left( \sum_{h\in [H]} \ell_{\infty, h} S\sqrt{AT\ln\left(\frac{SAT}{\delta}\right)} +\beta T  \sum_{h\in [H]} \ell_{\infty, h} \right).
\end{align*}
\end{lemma}

\noindent\textbf{Bounding \circled{2}}:
For simplicity, we denote by $q^s = \max_{\pi\in \Pi}q^{P,\pi}(s) $ and $\ell_t' = |\ell_t-\ell_t^c|$.
Notice that
\begin{align*}
    \sum_{t=1}^T \langle \ell_t-\ell_t^c, q_t-q^*\rangle 
    \le& \sum_{s\in [S], a\in [A]} \sum_{t=1}^T \ell_t'(s,a)|q_t(s,a)-q^*(s,a)|\\
    =& \sum_{s\in [S], a\in [A]} \sum_{t=1}^T \ell_t'(s,a)|q_t(s,a)-q^*(s,a)|
    \mathbbm{1}\left\{ q^s\le \mathcal{\tilde O}(SAH/\xi T)  \right\}\\
    &\quad+  \sum_{s\in [S], a\in [A]} \sum_{t=1}^T \ell_t'(s,a)|q_t(s,a)-q^*(s,a)|
    \mathbbm{1}\left\{ q^s> \mathcal{\tilde O}(SAH/\xi T)  \right\}
\end{align*}
For the first term, we can bound it directly by 
\begin{align*}
    &\sum_{s\in [S], a\in [A]} \sum_{t=1}^T \ell_t'(s,a)|q_t(s,a)-q^*(s,a)|
    \mathbbm{1}\left\{ q^s\le \mathcal{\tilde O}(SAH/\xi T)  \right\}\\
    &\le \sum_{h\in [H]}\ell_{\infty,h} \sum_{s\in [S_h]} 2 q^s T
     = \mathcal{\tilde O}\left(\frac{ \sum_{h\in [H]}\ell_{\infty,h}  S_h S A H}{\xi}  \right).
\end{align*}
This first inequality is due to $\sum_{ a\in [A]}\ell_t'(s,a)|q_t(s,a)-q^*(s,a)|\le 2\ell_{\infty,h(s)}q^s$ for all $s\in [S]$.

It suffices to focus on the second term.
By Lemma~\ref{lem:rfelp}, for every $s\in [S]$, if $q^s> \mathcal{\tilde O}(SAH/\xi T)$, $q^s$ and $q^{P,\pi^{s,\xi T}}$ will be on the same order.  
Thus, when 
\begin{align*}
    \sum_{ a\in [A]}\ell_t'(s,a)|q_t(s,a)-q^*(s,a)|
    \mathbbm{1}\left\{ q^s> \mathcal{\tilde O}(SAH/\xi T)  \right\}\not=0,
\end{align*}
the extra exploration policy ensures the outlier loss of $(s,a)$ has probability at least $\mathcal{\tilde O}(\beta q^s/SA) $ to be visited.
    Moreover, we can note that $\sum_{ a\in [A]}\ell_t'(s,a) |q_t(s,a)-q^*(s,a)|
   \le \mathcal{O}(q^s \ell_{\infty, h(s)} )$, and thus the terms dependent on $q^s$ can be eliminated.
   The result is presented below.
\begin{lemma}
\label{lemma:MDP_clip}
    With probability at least $1-\delta$,  
    \begin{align*}
        \sum_{s\in [S], a\in [A]} \sum_{t=1}^T \ell_t'(s,a)|q_t(s,a)-q^*(s,a)|
    \mathbbm{1}\bigg\{ q^s> \mathcal{\tilde O}&(SAH/\xi T)  \bigg\}\le \mathcal{\tilde O}\left(\frac{ \sum_{h\in [H]}\ell_{\infty,h}  SA}{\beta}  \right),
    \end{align*}
\end{lemma}
where the proof is in Appendix~\ref{appendix:lemma:MDP_clip}.
Summing up all the terms lead to regret
\begin{align*}
    \mathcal{R}(T)\le \mathcal{\tilde O}\bigg(&\sum_{h\in [H]}\ell_{\infty, h} \left[ S\sqrt{AT} +\beta T + \frac{   S_h S H A}{\xi} + \frac{   SA}{\beta}+  \xi S T \right]\bigg)
\end{align*}
Setting $\xi = \mathcal{O}(\sqrt{SA/T})$, $\beta = \mathcal{O}(\sqrt{SA/T})$ and $ S_h = S/H$ concludes the proof.

\section{Conclusion}
This paper initiates the study of scale-free learning in adversarial MDPs. Our framework \texttt{SCB} allows us to achieve the minimax optimal expected regret for scale-free adversarial MABs and the first known high-probability regret in both scale-free adversarial MAB and scale-free adversarial MDPs. Future work includes closing the gap in the $S$-dependency in the adversarial MDPs regret.



\acks{We thank a bunch of people and funding agency.}

\bibliography{sample}

\begin{thebibliography}{33}
\providecommand{\natexlab}[1]{#1}
\providecommand{\url}[1]{\texttt{#1}}
\expandafter\ifx\csname urlstyle\endcsname\relax
  \providecommand{\doi}[1]{doi: #1}\else
  \providecommand{\doi}{doi: \begingroup \urlstyle{rm}\Url}\fi

\bibitem[Abernethy and Rakhlin(2009)]{abernethy2009beating}
Jacob Abernethy and Alexander Rakhlin.
\newblock Beating the adaptive bandit with high probability.
\newblock In \emph{2009 Information Theory and Applications Workshop}, pages 280--289. IEEE, 2009.

\bibitem[Audibert et~al.(2009)Audibert, Bubeck, et~al.]{audibert2009minimax}
Jean-Yves Audibert, S{\'e}bastien Bubeck, et~al.
\newblock Minimax policies for adversarial and stochastic bandits.
\newblock In \emph{COLT}, volume~7, pages 1--122, 2009.

\bibitem[Auer et~al.(2002{\natexlab{a}})Auer, Cesa-Bianchi, and Fischer]{auer2002finite}
Peter Auer, Nicolo Cesa-Bianchi, and Paul Fischer.
\newblock Finite-time analysis of the multiarmed bandit problem.
\newblock \emph{Machine learning}, 47\penalty0 (2-3):\penalty0 235--256, 2002{\natexlab{a}}.

\bibitem[Auer et~al.(2002{\natexlab{b}})Auer, Cesa-Bianchi, Freund, and Schapire]{auer2002nonstochastic}
Peter Auer, Nicolo Cesa-Bianchi, Yoav Freund, and Robert~E Schapire.
\newblock The nonstochastic multiarmed bandit problem.
\newblock \emph{SIAM journal on computing}, 32\penalty0 (1):\penalty0 48--77, 2002{\natexlab{b}}.

\bibitem[Bubeck et~al.(2018)Bubeck, Cohen, and Li]{bubeck2018sparsity}
S{\'e}bastien Bubeck, Michael Cohen, and Yuanzhi Li.
\newblock Sparsity, variance and curvature in multi-armed bandits.
\newblock In \emph{Algorithmic Learning Theory}, pages 111--127. PMLR, 2018.

\bibitem[Cesa-Bianchi et~al.(2007)Cesa-Bianchi, Mansour, and Stoltz]{cesa2007improved}
Nicolo Cesa-Bianchi, Yishay Mansour, and Gilles Stoltz.
\newblock Improved second-order bounds for prediction with expert advice.
\newblock \emph{Machine Learning}, 66:\penalty0 321--352, 2007.

\bibitem[Chen and Zhang(2023)]{chen2023improved}
Mingyu Chen and Xuezhou Zhang.
\newblock Improved algorithms for adversarial bandits with unbounded losses.
\newblock \emph{arXiv preprint arXiv:2310.01756}, 2023.

\bibitem[Cutkosky(2019)]{cutkosky2019artificial}
Ashok Cutkosky.
\newblock Artificial constraints and hints for unbounded online learning.
\newblock In \emph{Conference on Learning Theory}, pages 874--894. PMLR, 2019.

\bibitem[Dai et~al.(2022)Dai, Luo, and Chen]{dai2022follow}
Yan Dai, Haipeng Luo, and Liyu Chen.
\newblock Follow-the-perturbed-leader for adversarial markov decision processes with bandit feedback.
\newblock \emph{Advances in Neural Information Processing Systems}, 35:\penalty0 11437--11449, 2022.

\bibitem[De~Rooij et~al.(2014)De~Rooij, Van~Erven, Gr{\"u}nwald, and Koolen]{de2014follow}
Steven De~Rooij, Tim Van~Erven, Peter~D Gr{\"u}nwald, and Wouter~M Koolen.
\newblock Follow the leader if you can, hedge if you must.
\newblock \emph{The Journal of Machine Learning Research}, 15\penalty0 (1):\penalty0 1281--1316, 2014.

\bibitem[Duchi et~al.(2011)Duchi, Hazan, and Singer]{duchi2011adaptive}
John Duchi, Elad Hazan, and Yoram Singer.
\newblock Adaptive subgradient methods for online learning and stochastic optimization.
\newblock \emph{Journal of machine learning research}, 12\penalty0 (7), 2011.

\bibitem[Even-Dar et~al.(2009)Even-Dar, Kakade, and Mansour]{even2009online}
Eyal Even-Dar, Sham~M Kakade, and Yishay Mansour.
\newblock Online markov decision processes.
\newblock \emph{Mathematics of Operations Research}, 34\penalty0 (3):\penalty0 726--736, 2009.

\bibitem[Freund and Schapire(1997)]{freund1997decision}
Yoav Freund and Robert~E Schapire.
\newblock A decision-theoretic generalization of on-line learning and an application to boosting.
\newblock \emph{Journal of computer and system sciences}, 55\penalty0 (1):\penalty0 119--139, 1997.

\bibitem[Hadiji and Stoltz(2023)]{hadiji2023adaptation}
H{\'e}di Hadiji and Gilles Stoltz.
\newblock Adaptation to the range in k--armed bandits.
\newblock \emph{Journal of Machine Learning Research}, 24\penalty0 (13):\penalty0 1--33, 2023.

\bibitem[Hazan and Kale(2011)]{hazan2011better}
Elad Hazan and Satyen Kale.
\newblock Better algorithms for benign bandits.
\newblock \emph{Journal of Machine Learning Research}, 12\penalty0 (4), 2011.

\bibitem[Huang et~al.(2023)Huang, Dai, and Huang]{huang2023banker}
Jiatai Huang, Yan Dai, and Longbo Huang.
\newblock Banker online mirror descent: A universal approach for delayed online bandit learning.
\newblock \emph{arXiv preprint arXiv:2301.10500}, 2023.

\bibitem[Ito(2021)]{ito2021parameter}
Shinji Ito.
\newblock Parameter-free multi-armed bandit algorithms with hybrid data-dependent regret bounds.
\newblock In \emph{Conference on Learning Theory}, pages 2552--2583. PMLR, 2021.

\bibitem[Jacobsen and Cutkosky(2023)]{jacobsen2023unconstrained}
Andrew Jacobsen and Ashok Cutkosky.
\newblock Unconstrained online learning with unbounded losses.
\newblock \emph{arXiv preprint arXiv:2306.04923}, 2023.

\bibitem[Jin et~al.(2019)Jin, Jin, Luo, Sra, and Yu]{jin2019learning}
Chi Jin, Tiancheng Jin, Haipeng Luo, Suvrit Sra, and Tiancheng Yu.
\newblock Learning adversarial mdps with bandit feedback and unknown transition.
\newblock \emph{arXiv preprint arXiv:1912.01192}, 2019.

\bibitem[Jin et~al.(2020)Jin, Krishnamurthy, Simchowitz, and Yu]{jin2020reward}
Chi Jin, Akshay Krishnamurthy, Max Simchowitz, and Tiancheng Yu.
\newblock Reward-free exploration for reinforcement learning.
\newblock In \emph{International Conference on Machine Learning}, pages 4870--4879. PMLR, 2020.

\bibitem[Jin et~al.(2022)Jin, Lancewicki, Luo, Mansour, and Rosenberg]{jin2022near}
Tiancheng Jin, Tal Lancewicki, Haipeng Luo, Yishay Mansour, and Aviv Rosenberg.
\newblock Near-optimal regret for adversarial mdp with delayed bandit feedback.
\newblock \emph{Advances in Neural Information Processing Systems}, 35:\penalty0 33469--33481, 2022.

\bibitem[Koc{\'a}k et~al.(2014)Koc{\'a}k, Neu, Valko, and Munos]{kocak2014efficient}
Tom{\'a}{\v{s}} Koc{\'a}k, Gergely Neu, Michal Valko, and R{\'e}mi Munos.
\newblock Efficient learning by implicit exploration in bandit problems with side observations.
\newblock \emph{Advances in Neural Information Processing Systems}, 27, 2014.

\bibitem[Lee et~al.(2020)Lee, Luo, Wei, and Zhang]{lee2020bias}
Chung-Wei Lee, Haipeng Luo, Chen-Yu Wei, and Mengxiao Zhang.
\newblock Bias no more: high-probability data-dependent regret bounds for adversarial bandits and mdps.
\newblock \emph{Advances in neural information processing systems}, 33:\penalty0 15522--15533, 2020.

\bibitem[Luo et~al.(2021)Luo, Wei, and Lee]{luo2021policy}
Haipeng Luo, Chen-Yu Wei, and Chung-Wei Lee.
\newblock Policy optimization in adversarial mdps: Improved exploration via dilated bonuses.
\newblock \emph{Advances in Neural Information Processing Systems}, 34:\penalty0 22931--22942, 2021.

\bibitem[Mayo et~al.(2022)Mayo, Hadiji, and van Erven]{mayo2022scale}
Jack~J Mayo, H{\'e}di Hadiji, and Tim van Erven.
\newblock Scale-free unconstrained online learning for curved losses.
\newblock In \emph{Conference on Learning Theory}, pages 4464--4497. PMLR, 2022.

\bibitem[Neu(2015)]{neu2015explore}
Gergely Neu.
\newblock Explore no more: Improved high-probability regret bounds for non-stochastic bandits.
\newblock \emph{Advances in Neural Information Processing Systems}, 28, 2015.

\bibitem[Orabona and P{\'a}l(2018)]{orabona2018scale}
Francesco Orabona and D{\'a}vid P{\'a}l.
\newblock Scale-free online learning.
\newblock \emph{Theoretical Computer Science}, 716:\penalty0 50--69, 2018.

\bibitem[Putta and Agrawal(2022)]{putta2022scale}
Sudeep~Raja Putta and Shipra Agrawal.
\newblock Scale-free adversarial multi armed bandits.
\newblock In \emph{International Conference on Algorithmic Learning Theory}, pages 910--930. PMLR, 2022.

\bibitem[Simchowitz and Jamieson(2019)]{simchowitz2019non}
Max Simchowitz and Kevin~G Jamieson.
\newblock Non-asymptotic gap-dependent regret bounds for tabular mdps.
\newblock \emph{Advances in Neural Information Processing Systems}, 32, 2019.

\bibitem[Talebi and Maillard(2018)]{talebi2018variance}
Mohammad~Sadegh Talebi and Odalric-Ambrym Maillard.
\newblock Variance-aware regret bounds for undiscounted reinforcement learning in mdps.
\newblock In \emph{Algorithmic Learning Theory}, pages 770--805. PMLR, 2018.

\bibitem[Wei and Luo(2018)]{wei2018more}
Chen-Yu Wei and Haipeng Luo.
\newblock More adaptive algorithms for adversarial bandits.
\newblock In \emph{Conference On Learning Theory}, pages 1263--1291. PMLR, 2018.

\bibitem[Zanette and Brunskill(2019)]{zanette2019tighter}
Andrea Zanette and Emma Brunskill.
\newblock Tighter problem-dependent regret bounds in reinforcement learning without domain knowledge using value function bounds.
\newblock In \emph{Proceedings of the 36th International Conference on Machine Learning}, 2019.

\bibitem[Zhang et~al.(2023)Zhang, Chen, Lee, and Du]{zhang2023settling}
Zihan Zhang, Yuxin Chen, Jason~D Lee, and Simon~S Du.
\newblock Settling the sample complexity of online reinforcement learning.
\newblock \emph{arXiv preprint arXiv:2307.13586}, 2023.

\end{thebibliography}

\appendix


\newpage

\section{Omitted details for Section 3}
\subsection{Proof of Lemma~\ref{lem:minimax_1}}
\label{appendix:lem:minimax_1}
We start by a technical lemma inspired by \cite{chen2023improved}. 
\begin{lemma} 
\label{Lemma::FTRL_regret}
For any $\hat \ell_1,\dots,\hat \ell_T\ge 0 $, using the update rule of (\ref{FTRL::update_rule}), consider any convex regularizer $\Psi\ge 0$ that satisfies $\nabla_{k,k} \Psi(\p)\le \nabla_{k,k} \Psi(\q)$ iff. $p_{k}\ge q_{k}$ for every $k\in [n]$ and $\p,\q\in \Delta_n$.
With non-increasing sequence of learning rates $\eta_1,\dots,\eta_{T+1}$, there is
    \begin{align*}
    \sum_{t=1}^T \langle \hat \ell_t, \p_t-\p^\dagger \rangle\le \frac{\Psi(\p^\dagger) }{\eta_{T+1}}+ \frac{1}{2} \sum_{t=1}^T \eta_{t}\| \hat \ell_t\|^2_{(\nabla^2 
 \Psi(\p_t))^{-1}}
\end{align*}
for every comparator $\p^\dagger \in \Delta_n$.
\end{lemma}

Recall the definition of $1/2$-Tsallis Entropy 
\begin{align*}
    \Psi(\p_t) = 4\sqrt{n}-4\sum_{k=1}^n \sqrt{p_{t,k}}.
\end{align*}
Notice that $\Psi(\p)\ge 0$ for every $\p\in \Delta_n$ and $\nabla_{k,k} \Psi(\p) = p_{k}^{-3/2}\le q_{k}^{-3/2} =  \nabla_{k,k} \Psi(\q)$ when $p_k\ge q_k$.
Using Lemma~\ref{Lemma::FTRL_regret}, there is
\begin{align*}
    \sum_{t=1}^T \langle \hat \ell_t, \p_t-\p^\star \rangle
    &\le \frac{\Psi(\p^\star) }{\eta_{T+1}}+ \frac{1}{2} \sum_{t=1}^T\eta_{t}\| \hat \ell_t\|^2_{(\nabla^2 \Psi(\p_t))^{-1}}\\
    &\le \frac{\Psi(\p^\star) }{\eta_{T+1}} + \frac{1}{2} \sum_{t=1}^T \eta_{t} \frac{({\ell}_{t,k_t}^c+C_t)^2}{q_{t,k_t}^2} p_{t,k_t}^{3/2} \\
    &\le \frac{\Psi(\p^\star) }{\eta_{T+1}} + 8 \sum_{t=1}^T \eta_{t} \frac{({\ell}_{t,k_t}^c+C_t)^2}{\sqrt{q_{t,k_t}}}
\end{align*}
The last inequality is due to $p_{t,k_t}\le  q_{t,k_t}/(1-\beta_t) \le  q_{t,k_t}/(1-n/(2n+\sqrt{nt}))\le 2q_{t,k_t}$.
We further note that, by the clipping rule, there is $|\ell_{t,k_t}^c+C_t|\le 2 C_t$.
Moreover, it suffices to say that $C_{t+1}\le 2\ell_\infty$ for all $t\in [T]$.
Remind $\Psi(\p)\le \sqrt{n}$ for every $\p \in \Delta_n$.
Thus, by choosing learning rate $\eta_t = {1}/{2 C_t\sqrt{t}}$,
we have
\begin{align*}
    \E\left[\sum_{t=1}^T \langle \hat \ell_t, \p_t-\p^\star \rangle\right] &\le \frac{\Psi(\p^\star) }{\eta_{T+1}}+ 8 \sum_{t=1}^T \E\left[ \eta_{t} \frac{({\ell}_{t,k_t}^c+C_t)^2}{\sqrt{q_{t,k_t}}}\right]\\
    &\le \frac{\Psi(\p^\star) }{\eta_{T+1}}+8 \sum_{t=1}^T\frac{1}{\sqrt{t}}\E\left[ \frac{|{\ell}_{t,k_t}^c+C_t|}{\sqrt{q_{t,k_t}}}\right]\\
    &\le 4\ell_\infty \sqrt{n (T+1)}+ 16\ell_\infty \sqrt{n}(2\sqrt{T}+1) \\
    &= \Theta(\ell_\infty \sqrt{nT}),
\end{align*}
where the third inequality is due to $\E[1/\sqrt{q_{t,k_t}}] = \sum_{k=1}^n \sqrt{q_{t,k_t}}\le \sqrt{n}$.

\subsection{Omitted details of Remark~\ref{remark:1}}
\label{apendix:remark:1}
Here we prove that \texttt{SCB} is invariant to rescaling of losses.
Notice that the proof also applies to \texttt{SCB-IX} and \texttt{SCB-RL}.
Starting with losses $\ell_1,\dots,\ell_T$, the rescaled losses are defined by $\ell_1',\dots, \ell_T'$ such that $\ell_t = c \ell_t'$ for all $t\in [T]$ and $c>0$.
With the use of \texttt{SCB}, the action distributions corresponding to these two sequences of losses are represented by $\p_1,\dots, \p_T$ and $\p_1',\dots, \p_T'$, respectively.
Notice that $\p_t$ can be considered as a random vector w.r.t. \texttt{SCB}, losses $\ell_1,\dots, \ell_{t-1}$ and past actions $k_1,\dots, k_t$. 
Our goal is to prove that the distributions of $\p_1,\dots, \p_T$ and $\p_1',\dots, \p_T'$ are the same. 
We prove by induction.
For $t=1$, both $\p_1$ and $\p_1'$ are uniform distribution over $[n]$.
Assuming at time $t$, the distributions of $\p_1,\dots, \p_t$ and $\p_1',\dots, \p_t'$ are the same.
Conditioned on $\p_1,\dots, \p_t = \p_1',\dots, \p_t'$, since $\beta_1,\dots,\beta_t$ are independent to the losses, for \texttt{SCB} with these two loss sequences, the probability of taking actions $\{k_1,\dots, k_{t}\}$ is the same.
Conditioned on actions $\{k_1,\dots, k_{t}\}$, we have $\hat \ell_s = c\hat\ell_s'$ for all $s\le t$.
Then, since the clipping threshold is twice the largest scale among the previously observed
losses, we further have $C_{t+1} = c C_{t+1}'$.
Thus, by the update rule, it suffices to show that $\p_{t+1} = \p_{t+1}'$.
This implies that the distributions of $\p_1,\dots, \p_{t+1}$ and $\p_1',\dots, \p_{t+1}'$ are the same, which completes the proof.

We emphasize that the use of clipping (or skipping) to deal with unbounded losses has been studied before \citep{chen2023improved, huang2023banker}.
However, our algorithms fundamentally differ from the previous ones. 
In previous works, the update of the clipping threshold is accomplished through a double trick, i.e., \(C_{t+1} = 2C_t\). 
This leads to an inevitable logarithm sub-optimality.
More importantly, their algorithms must start from a positive clipping threshold \(C_1 > 0\), resulting in a failure to achieve strongly scale-free.
Relatively, our algorithm starts from \(C_1 = 0\). 
This allows the clipping threshold to be linearly related to the scale of the losses, thereby achieving strongly scale-free.

\subsection{Proof of Theorem~\ref{theo:highprop}}
\label{appendix: high_prop}
By Hoeffding's inequality, there is 
\begin{align*}
        \sum_{t=1}^T \ell_{t,k_t} - \sum_{t=1}^T \ell_{t, k^\star} \le \ell_\infty\sqrt{2T\log(1/\delta)}+\sum_{t=1}^T \langle \ell_t, \q_t-\p^\star \rangle
    \end{align*}
with probability at least $1-\delta$.
It suffices to focus on bounding $\sum_{t=1}^T \langle \ell_t, \q_t-\p^\star \rangle$.
Similar to the proof of Theorem~\ref{theo:mmopt}, we decompose the regret into  
\begin{align*}
    \sum_{t=1}^T \langle \ell_t, \q_t-\p^\star \rangle = \underbrace{\sum_{t=1}^T \langle \ell_t^c+C_t\mathbf{1}_n, \q_t-\p^\star  \rangle}_{\circled{1}}  +\underbrace{\sum_{t=1}^T \langle \ell_t - \ell_t^c, \q_t-\p^\star  \rangle}_{\circled{2}}
\end{align*}
and bound these two terms respectively.

\noindent\textbf{Bounding $\circled{1}$}:
The high level idea of bounding $\circled{1}$ is inspired by the proof of Theorem $1$ in \cite{neu2015explore}.
Compared to \texttt{EXP3-IX}, our algorithm adds an additional explicit exploration, i.e., mixes $\p_t$ with uniform distribution, and $\ell_{t,k}^c+C_t$ is within $[0,2C_t]$ instead of $[0,1]$.
Using a similar idea proposed in the previous section, we can show that \circled{1} can be well bounded with high probability.
The detailed proof is provided in the appendix.
\begin{lemma}
\label{High_prop:1}
    With probability at least $1-\delta$, there is 
    \begin{align*}
        \sum_{t=1}^T \langle \ell_t^c+C_t\mathbf{1}_n, \q_t-\p^\star  \rangle \le \Theta\left( \ell_\infty   \sqrt{\frac{nT}{\log n}}\log(1/\delta)  + \ell_\infty\sqrt{ nT \log (n)}\right).
    \end{align*}
\end{lemma}

\noindent\textbf{Bounding \circled{2}}: 
Define $K:=\arg\min_{j\in \mathbb{N}}\left\{  \ell_\infty \le 2^j \right\}$.
Define $\ell_t^i\in \R^n$ such that $\ell_{t,k}^i = \ell_{t,k}\mathbbm{1}\{2^{i-1}<\ell_{t,k}\le 2^i\}$ for $k\in [n]$.
Inspired by the results of the above section, we note that the clipping error can be reduced to the sum of the error incurred by losses within $[2^{i-1}, 2^i]$, i.e.,
\begin{align*}
    \sum_{t=1}^T \langle \ell_t-\ell_t^c, \p_t-\p^\dagger   \rangle \le 2 \sum_{i=-\infty}^K\sum_{t=1}^{T} \|\ell_t^i - {\ell_t^c}^i\|_\infty \le 2 \sum_{i=-\infty}^K 2^i \left( \sum_{t=1}^{T} \mathbbm{1}\{\ell_t^i\not=\mathbf{0}_n\}  \mathbbm{1}\{C_t<2^i\} \right)
\end{align*}
Apparently, bounding $\sum_{t=1}^{T} \mathbbm{1}\{\ell_t^i\not=\mathbf{0}_n\}  \mathbbm{1}\{C_t<2^i\}$ individually is not difficult due to
\begin{align*}
    \P\left\{\sum_{t=1}^{T} \mathbbm{1}\{\ell_t^i\not=\mathbf{0}_n\}  \mathbbm{1}\{C_t<2^i\}> \frac{n}{\beta_t}\log(1/\delta)  \right\}\le \left(1-\frac{\beta_t}{n}\right)^{\frac{n}{\beta_t}\log(1/\delta)} \le \delta.
\end{align*}
The challenge is how to achieve a union bound on all $i\le K$ without losing any optimality of the logarithmic terms.
This is shown in the following lemma. 
\begin{lemma}
\label{High_prop:2}
    With probability at least $1-\delta$,
    \begin{align*}
        \sum_{t=1}^T \langle \ell_t-\ell_t^c, \p_t-\p^\star   \rangle \le \Theta\left(\ell_\infty    \sqrt{\frac{n^2+nT}{\log n} } \log(1/\delta)\right)
    \end{align*}
\end{lemma}
Given Lemma~\ref{High_prop:1} and Lemma~\ref{High_prop:2}, we can obtain the results of Theorem~\ref{theo:highprop}.

\subsection{Proof of Lemma~\ref{Lemma::FTRL_regret}}
The proof refers to Lemma $1$ and $2$ in \cite{chen2023improved}.
Define 
\begin{align*}
    F_t(\p) = \sum_{s=1}^{t-1} \langle \hat\ell_s, \p \rangle +\frac{1}{\eta_t}\Psi(\p),
\end{align*}
we first note that
\begin{align*}
    \sum_{t=1}^T \langle \hat \ell_t, \p_t-\p^\dagger \rangle &=   -F_{T+1}(\p^\dagger)+\frac{1}{\eta_{T+1}}\Psi(\p^\dagger)+\sum_{t=1}^T \langle \hat \ell_t, \p_t \rangle\\
    &=-F_{T+1}(\p^\dagger)+\frac{1}{\eta_{T+1}}\Psi(\p^\dagger)-F_1(\p_1)+F_{T+1}(\p_{T+1}) \nonumber \\  
    &+\sum_{t=1}^T \left( F_t(\p_t)-F_{t+1}(\p_{t+1}) \right)+\sum_{t=1}^T \langle \hat \ell_t, \p_t \rangle\\
    &=-F_{T+1}(\p^\dagger)+\frac{1}{\eta_{T+1}}\Psi(\p^\dagger)-F_1(\p_1)+F_{T+1}(\p_{T+1}) \nonumber \\  
    &+\sum_{t=1}^T \left( F_t(\p_t)+\langle \hat \ell_t, \p_t \rangle-F_{t+1}(\p_{t+1}) \right).
\end{align*}
By definition, there is 
\begin{align*}
    F_{T+1}(\p_{T+1})-F_{T+1}(\p^\dagger)&=\min_{\p\in \Delta_n}F_{T+1}(\p)-F_{T+1}(\p^\dagger) \le 0\\
    \frac{1}{\eta_{T+1}}\Psi(\p^\dagger)-F_1(\p_1) &= \frac{1}{\eta_{T+1}}\Psi(\p^\dagger)- \min_{\p\in \Delta_n }\Psi_{1}(\p)\le  \frac{1}{\eta_{T+1}}\Psi(\p^\dagger).
\end{align*}
Thus, we obtain 
\begin{align*}
    \sum_{t=1}^T \langle \hat \ell_t, \p_t-\p^\dagger \rangle\le \frac{1}{\eta_{T+1}}\Psi(\p^\dagger) + \sum_{t=1}^T \left( F_t(\p_t)+\langle \hat \ell_t, \p_t \rangle-F_{t+1}(\p_{t+1}) \right)
\end{align*}
Furthermore, we note that
\begin{align*}
     F_t(\p_t)+\langle \hat \ell_t, \p_t \rangle-F_{t+1}(\p_{t+1})&=\sum_{s=1}^{t}\langle \hat \ell_s, \p_t-\p_{t+1} \rangle + \frac{1}{\eta_{t}}\Psi(\p_t)- \frac{1}{\eta_{t+1}}\Psi(\p_t) \\
     &\le \sum_{s=1}^{t}\langle \hat \ell_s, \p_t-\p_{t+1} \rangle + \frac{1}{\eta_{t}}\Psi(\p_t)- \frac{1}{\eta_{t}}\Psi(\p_t)\\
     &=\langle \hat \ell_t, \p_t-\p_{t+1} \rangle+ F_t(\p_t)-F_{t}(\p_{t+1}),
\end{align*}
where the first inequality is due to the assumption $\eta_{t+1}\le \eta_{t}$.
By Taylor’s expansion, we have
\begin{align*}
    F_t(\p_{t+1})-F_t(\p_t) = \langle \nabla F_t(\p_t), \p_{t+1}-\p_t \rangle+\frac{1}{2}\|\p_{t+1}-\p_t\|^2_{\nabla^2 F_t(\xi_t)}.
\end{align*}
where $\xi_t=\alpha \p_t+(1-\alpha)\p_{t+1}$ for some $\alpha\in [0,1]$.
By definition,
\begin{align*}
    \p_t =\arg\min_{\p\in \Delta_n} F_t(\p).
\end{align*}
By KKT conditions, there exists some $\lambda_t\in \R$ such that
\begin{align*}
     \p_t =\arg\min_{\p\in \R} \Big(F_t(\p) +\lambda_t (1-\sum_{k=1}^n p_{t,k})\Big).
\end{align*}
By the optimality of $\p_t$, we have
\begin{align*}
    \nabla F_t(\p_t) +\lambda_t\mathbf{1}_n = 0,
\end{align*}
which implies 
\begin{align*}
    \langle \nabla F_t(\p), \p_{t+1}-\p_t \rangle =  \langle -\lambda_t\mathbf{1}_n, \p_{t+1}-\p_t \rangle = 0. 
\end{align*}
Thus, there is 
\begin{align*}
    F_t(\p_{t+1})-F_t(\p_t) &= \frac{1}{2}\|\p_{t+1}-\p_t\|^2_{\nabla^2 F_t(\xi_t)}.
\end{align*}
Using the above, 
\begin{align*}
     \langle \hat \ell_t, \p_t-\p_{t+1} \rangle+ F_t(\p_t)-F_{t}(\p_{t+1})&= \langle \hat \ell_t, \p_t-\p_{t+1} \rangle-\frac{1}{2}\|\p_{t+1}-\p_t\|^2_{\nabla^2 F_t(\xi_t)}  \\
    & \le \max_{\p\in \R} \Big( \langle \hat \ell_t, \p \rangle-\frac{1}{2}\|\p\|^2_{\nabla^2 F_t(\xi_t)} \Big) \\
    &\le \frac{1}{2}\|\hat \ell_t\|^2_{(\nabla^2 F_t(\xi_t))^{-1}} = \frac{1}{2}\eta_{t}\|\hat \ell_t\|^2_{(\nabla^2 \Psi(\xi_t))^{-1}},
\end{align*}
where the second inequality is because $\nabla^2 \Psi(\xi_t)$ is a diagonal matrix and the second equality is due to $\nabla^2 F_t(\xi_t) = \nabla^2 \Psi(\xi_t)/\eta_{t} $.
Now we prove
\begin{align*}
    \langle \hat \ell_t, \p_t-\p_{t+1} \rangle+ F_t(\p_t)-F_{t}(\p_{t+1})\le \frac{1}{2}\eta_{t}\| \hat \ell_t\|^2_{(\nabla^2 
 \Psi(\p_t))^{-1}}
\end{align*}
if $\hat\ell_t\in \R_+^n$.
Recall
\begin{align*}
     \| \hat \ell_t\|^2_{(\nabla^2  \Psi(\xi_t))^{-1}} = \sum_{k=1}^n \frac{\hat \ell_{t,k}^2 }{\nabla^2_{k,k}  \Psi(\xi_t)}  =  \frac{\hat \ell_{t,k_t}^2 }{\nabla^2_{k_t,k_t}  \Psi(\xi_t)}  
\end{align*}
and $\xi_t$ is between $\p_t$ and $\p_{t+1}$, we prove case by case.
\begin{enumerate}
    \item ($ p_{t,k_t}- p_{t+1,k_t}<0$): In this case, we have 
    \begin{align*}
        \langle \hat \ell_t, \p_t-\p_{t+1} \rangle+ F_t(\p_t)-F_{t}(\p_{t+1})&\le \langle \hat \ell_t, \p_t-\p_{t+1} \rangle\\
        &= \hat \ell_{t,k_t}(p_{t,k_t}- p_{t+1,k_t})\\
        &\le 0 \le \frac{1}{2} \| \hat \ell_t\|^2_{(\nabla^2  \Psi(\p_t))^{-1}}.
    \end{align*}
    The first inequality is due to $\p_t$ minimizing $F_t$.
    The second inequality is due to $\hat\ell_{t,k_t}\ge 0$.
    \item ($ p_{t,k_t}- p_{t+1,k_t}\ge 0$): 
    In this case, we have $ p_{t,k_t}\ge \xi_{t,k_t}$.
    By assumption, there is $\nabla^2_{k_t,k_t}  \Psi(\p_t)\le \nabla^2_{k_t,k_t}  \Psi(\xi_t)$.  
    Thus
    \begin{align*}
        \| \hat \ell_t\|^2_{(\nabla^2  \Psi(\xi_t))^{-1}}=  \frac{\hat \ell_{t,k_t}^2 }{\nabla^2_{k_t,k_t}  \Psi(\xi_t)}\le \frac{\hat \ell_{t,k_t}^2 }{\nabla^2_{k_t,k_t}  \Psi(\p_t)} = \| \hat \ell_t\|^2_{(\nabla^2  \Psi(\p_t))^{-1}}
    \end{align*}
    completes the proof.
\end{enumerate}

\subsection{Proof of Lemma~\ref{High_prop:1}}
 We start by introducing a concentration result of the implicit exploration estimator based on Lemma $1$ in \cite{neu2015explore}.
\begin{lemma}
\label{High_prop:cl}
    Let $\gamma_1,\dots, \gamma_T$ be a fixed non-increasing sequence with $\gamma_t\ge 0,\ \forall t\in [T]$ and $\alpha_{t,k}$ be non-negative $\mathcal{F}_{t-1}$ measurable random variables satisfying $\alpha_{t,k}\le 2\gamma_t,\ \forall t\in [T],k\in [n]$. Then, with probability at least $1-\delta$, 
    \begin{align*}
        \sum_{t=1}^T\sum_{k=1}^n \alpha_{t,k}\left(\hat\ell_{t,k} - (\ell_{t,k}^c+C_t) \right)\le 3\ell_\infty \log(1/\delta).
    \end{align*}
\end{lemma}

Given Lemma~\ref{High_prop:cl}, we decompose the \circled{1} into $4$ terms.
\begin{align*}
    &\sum_{t=1}^T \langle \ell_{t}^c+C_t\mathbf{1}_n, \q_t-\p^\dagger  \rangle = \sum_{t=1}^T \langle \hat \ell_t, \p_t-\p^\dagger  \rangle + \sum_{t=1}^T \langle \hat \ell_t, \q_t-\p_t  \rangle + \sum_{t=1}^T \langle \ell_{t}^c+C_t\mathbf{1}_n -\hat \ell_t, \q_t-\p^\dagger  \rangle\\
    & = \sum_{t=1}^T \langle \hat \ell_t, \p_t-\p^\dagger  \rangle + \sum_{t=1}^T \langle \hat \ell_t, \q_t-\p_t  \rangle + \sum_{t=1}^T \langle \ell_{t}^c+C_t\mathbf{1}_n -\hat \ell_t, \q_t  \rangle + \sum_{t=1}^T \langle \hat \ell_t - (\ell_{t}^c+C_t\mathbf{1}_n), \p^\dagger  \rangle.
\end{align*}

For the first term, recall the definition of (negative) Shannon Entropy 
\begin{align*}
    \Psi(\p_t) = \log n + \sum_{k=1}^n p_{t,k} \log(p_{t,k}).
\end{align*}
Notice that $ 0\le \Psi(\p)\le \log n$ for every $\p\in \Delta_n$ and $\nabla_{k,k} \Psi(\p) = p_{k}^{-1}\le q_{k}^{-1} =  \nabla_{k,k} \Psi(\q)$ when $p_k\ge q_k$.
Using Lemma~\ref{Lemma::FTRL_regret}, there is 
\begin{align*}
    \sum_{t=1}^T \langle \hat \ell_t, \p_t-\p^\dagger  \rangle&\le \frac{\log n}{\eta_{T+1}}+\frac{1}{2} \sum_{t=1}^T \eta_t p_{t,k_t}  (\hat \ell_{t,k_t})^2\\
    &\le \frac{\log n}{\eta_{T+1}}+ 2\sum_{t=1}^T  \eta_t C_t \hat \ell_{t,k_t}\\
    &\le  3\ell_\infty \sqrt{n(T+1)\log (n) } + 4\sum_{t=1}^T  \gamma_t\hat \ell_{t,k_t}.
\end{align*}
The second inequality is due to $p_{t,k_t}\hat \ell_{t,k_t} = p_{t,k_t}( \ell_{t,k_t}^c+C_t)/(q_{t,k_t}+\gamma_t)\le 4C_t$.
Since $\eta_t C_t \le 2\gamma_t$, using Lemma~\ref{High_prop:cl} and setting $\alpha_{t,k} = \gamma_t$ for every $k\in [n]$, we have
\begin{align*}
    \sum_{t=1}^T  \gamma_t \hat \ell_{t,k_t}&\le\sum_{t=1}^T  \gamma_t  \sum_{k=1}^n (\ell_{t,k}^c+C_t)+3\ell_\infty {\log(1/\delta)}\\
    &\le  3\ell_\infty \sqrt{\log (n) } \sum_{t=1}^T \sqrt{\frac{n}{t}}+3\ell_\infty {\log(1/\delta)}\\
    &\le 6 \ell_\infty \sqrt{ n T \log (n) } + 3\ell_\infty {\log(1/\delta)}.
\end{align*}
The second inequality is due to $\ell_{t,k}^c+C_t\le 3\ell_\infty$ for all $t\in [T]$.
Combining the above, there is 
\begin{align*}
    \sum_{t=1}^T \langle \hat \ell_t, \p_t-\p^\dagger  \rangle\le \Theta\left(\ell_\infty \sqrt{nT \log (n)} +\ell_\infty \log (1/\delta) \right)
\end{align*}
with probability at least $1-\delta$.

For the second term, we note that
\begin{align*}
    \sum_{t=1}^T \langle \hat \ell_t, \q_t-\p_t  \rangle &\le \sum_{t=1}^T \beta_t \left\langle \hat \ell_t, \frac{\mathbf{1}_n}{n}  \right\rangle\\
     & \le  \sum_{t=1}^T \beta_t \left\langle \hat \ell_t-(\ell_t^c+C_t\mathbf{1}_n), \frac{\mathbf{1}_n}{n}  \right\rangle + \sum_{t=1}^T \beta_t \left\langle \ell_t^c+C_t\mathbf{1}_n, \frac{\mathbf{1}_n}{n}  \right\rangle \\
    & \le  \sum_{t=1}^T \beta_t \left\langle \hat \ell_t-(\ell_t^c+C_t\mathbf{1}_n), \frac{\mathbf{1}_n}{n}  \right\rangle + 6\ell_\infty\sqrt{nT \log (n)}.
\end{align*}
Since $\beta_t/n \le 2\gamma_t$, using Lemma~\ref{High_prop:cl} and setting $\alpha_{t,k} = \beta_t /n$ for every $k\in [n]$, we have
\begin{align*}
    \sum_{t=1}^T \beta_t \left\langle \hat \ell_t-(\ell_t^c+C_t\mathbf{1}_n), \frac{\mathbf{1}_n}{n}  \right\rangle \le \sum_{t=1}^T 2\gamma_t \sum_{k=1}^n  \left(\hat\ell_{t,k}-(\ell_{t,k}^c+C_t) \right)\le 3\ell_\infty\log(1/\delta).
\end{align*}
Thus we have 
\begin{align*}
    \sum_{t=1}^T \langle \hat \ell_t, \q_t-\p_t  \rangle\le  \Theta\left( \ell_\infty\sqrt{nT \log (n)} + \ell_\infty\log(1/\delta) \right)
\end{align*}
with probability at least $1-\delta$.

For the third term, there is
\begin{align*}
    \sum_{t=1}^T \langle \ell_t^c+C_t\mathbf{1}_n -\hat \ell_t, \q_t  \rangle 
    &= \sum_{t=1}^T \sum_{k=1}^n  q_{t,k}(\ell_{t,k}^c+C_t -\hat \ell_{t,k}) \\
    & = \sum_{t=1}^T \sum_{k=1}^n  q_{t,k}\left(\ell_{t,k}^c+C_t-\frac{\mathbbm{1}\{k=k_t\}}{ q_{t,k}+\gamma_t}(\ell_{t,k}^c+C_t)\right) \\
    & = \sum_{t=1}^T \sum_{k=1}^n  q_{t,k}\left(\ell_{t,k}^c+C_t-\frac{q_{t,k}}{ q_{t,k}+\gamma_t}(\ell_{t,k}^c+C_t)\right)\\
    &+\sum_{t=1}^T \sum_{k=1}^n  q_{t,k}\left(\frac{q_{t,k}}{ q_{t,k}+\gamma_t}(\ell_{t,k}^c+C_t)-\frac{\mathbbm{1}\{k=k_t\}}{ q_{t,k}+\gamma_t}(\ell_{t,k}^c+C_t)\right) \\
    &\le \sum_{t=1}^T \sum_{k=1}^n \gamma_t(\ell_{t,k}^c+C_t)+\sum_{t=1}^T \langle \widetilde{\ell}_t,    \q_t-\mathbf{e}_{k_t} \rangle,
\end{align*}
where $\widetilde{\ell_t}$ denotes the implicit loss vector such that $\widetilde{\ell}_{t,k} = \frac{q_{t,k}}{ q_{t,k}+\gamma_t}(\ell_{t,k}^c+C_t)$ for every $k\in [n]$.
Notice that $\|\widetilde{\ell_t}\|_\infty\le 3\ell_\infty$.
Thus with probability at least $1-\delta$,
\begin{align*}
    \sum_{t=1}^T \sum_{k=1}^n \gamma_t(\ell_{t,k}^c+C_t)+\sum_{t=1}^T \langle \widetilde{\ell}_t,    \q_t-\mathbf{e}_{k_t} \rangle &\le 6\ell_\infty\sqrt{nT\log (n)} + 3\ell_\infty \sqrt{2T\log (2/\delta)}\\
    &\le \Theta\left(\ell_\infty\sqrt{nT\log (n)}+ \ell_\infty\sqrt{T\log(1/\delta)} \right).
\end{align*}

For the last term, 
\begin{align*}
    \sum_{t=1}^T \langle \hat \ell_t - (\ell_t^c+C_t\mathbf{1}_n), \p^\dagger  \rangle  &\le \frac{1}{2 \gamma_T} \sum_{t=1}^T 2\gamma_t \langle \hat \ell_t - (\ell_t^c+C_t\mathbf{1}_n), \p^\dagger  \rangle\\
    &\le \frac{3}{2 \gamma_T}\ell_\infty  \log(1/\delta)\\
    &\le \Theta\left( \ell_\infty   \sqrt{\frac{nT}{\log n}}\log(1/\delta)   \right).
\end{align*}
Summing up the above we can bound \circled{1} by $\Theta\left( \ell_\infty   \sqrt{\frac{nT}{\log n}}\log(1/\delta)  + \ell_\infty\sqrt{ nT \log (n)}\right)$. 

\subsection{Proof of Lemma~\ref{High_prop:2}}
Remind 
\begin{align*}
    \P\left\{\sum_{t=1}^{T} \mathbbm{1}\{\ell_t^i\not=\mathbf{0}_n\}  \mathbbm{1}\{C_t<2^i\}> \frac{n}{\beta_t}\log(1/\delta)  \right\}  \le \delta
\end{align*}
for every $i\le K$.
We first note that
\begin{align*}
     \sum_{i=-\infty}^K 2^i \left( \sum_{t=1}^{T} \mathbbm{1}\{\ell_t^i\not=\mathbf{0}_n\}  \mathbbm{1}\{C_t<2^i\} \right)\le  \sum_{i=-\infty}^K 2^i \left(   \sqrt{\frac{n^2+nT}{\log n} } \log(2^{K-i+2}/\delta) \right)
\end{align*}
with probability at least $1-\delta$. 
This is because $\sum_{i=-\infty}^K \delta/2^{K-i+2}\le \delta$.
Denote by 
\begin{align*}
    S_i = 2^i \left(   \sqrt{\frac{n^2+nT}{\log n} } \log(2^{K-i+2}/\delta) \right),
\end{align*}
we then prove $S_{i-1}/S_i\le 3 /4$ for every $i\le K$.
\begin{align*}
    \frac{S_{i-1}}{S_i} \le \frac{1}{2} \frac{\log(2^{K-i+2}/\delta)+\log 2}{\log(2^{K-i+2}/\delta)}\le \frac{3}{4}.
\end{align*}
Thus,
\begin{align*}
    \sum_{i=-\infty}^K 2^i \left( \sum_{t=1}^{T} \mathbbm{1}\{\ell_t^i\not=\mathbf{0}_n\}  \mathbbm{1}\{C_t<2^i\} \right)&\le \sum_{i=-\infty}^K S_i\le S_K \sum_{i=-\infty}^K (\frac{3}{4})^{K-i} = 4 S_K\\
    &\le 2^{K+2} \left(   \sqrt{\frac{n^2+nT}{\log n} } \log(4/\delta) \right)\\
    &\le 8\ell_\infty    \sqrt{\frac{n^2+nT}{\log n} } \log(4/\delta) \\
    &= \Theta\left(\ell_\infty    \sqrt{\frac{n^2+nT}{\log n} } \log(1/\delta)\right)
\end{align*}

\subsection{Proof of Lemma~\ref{High_prop:cl}}
Notice that
    \begin{align*}
        \sum_{t=1}^T\sum_{k=1}^n \alpha_{t,k}\left(\hat\ell_{t,k} - (\ell_{t,k}^c+C_t) \right)= 3\ell_\infty \sum_{t=1}^T\sum_{k=1}^n \alpha_{t,k}\left( 
\frac{\hat\ell_{t,k}}{3\ell_\infty} - \frac{\ell_{t,k}^c+C_t}{3\ell_\infty} \right).
    \end{align*}
Since $0\le \ell_{t,k}^c+C_t\le 3\ell_\infty$ for all $t\in [T]$, by Lemma 1 in \cite{neu2015explore}, we complete the proof. 
    
\section{Omitted details for Section 4}

\subsection{Omitted details of Occupancy measure}
In this subsection, we briefly explain the concept of ``occupancy measure'' and show how to reformulate adversarial MDP problems to adversarial MAB problems (for more details see \cite{jin2019learning} and \cite{lee2020bias}).
For any $(s,a)\in [S]\times[A]$, the probability that policy $\pi$ visits the state-action pair $(s,a)$ with transition function $P$ can be denoted by  
\begin{align*}
    q^{P, \pi}(s,a) = \P\left\{ s_{h(s)} = s, a_{h(s)} = a    |P, \pi \right\},
\end{align*}
where $h(s)$ denotes the index of the layer to which state $s$ belongs.
Here, $q^{P,\pi}\in R^{S\times A}$ is a valid occupancy measure.
Following \cite{jin2019learning}, we denote $\Delta(P)$ by the set of occupancy measures whose induced transition function is $P$, i.e., the set of $q^{P,\pi}$ for all policy $\pi$ with transition function $P$, and $\Delta(\mathcal{P})$ by the set of occupancy measures whose induced transition function belongs to the set of transition functions $\mathcal{P}$, i.e., the set of $q^{P,\pi}$ for all policy $\pi$ with transition function $P\in \mathcal{P}$.
Assuming $P$ is the underlying transition function, the total expected regret (w.r.t. randomness of the transition function) can be written as 
\begin{align*}
 \mathcal{R}(T) &= \sum_{t=1}^T \ell_t(\pi_t)- \sum_{t=1}^T \ell_t(\pi^\star) \\
 &= \sum_{t=1}^T \sum_{s\in [S], a\in [A]}(q^{P,\pi_t}(s,a)-q^{P,\pi^\star}(s,a))\ell_t(s,a) \\&=\sum_{t=1}^T \langle q^{P,\pi_t}- q^{P,\pi^\star}, \ell_t   \rangle.
\end{align*}
When the regret is written in this way, it is clear that the adversarial MDP problems can be reduced to the adversarial MAB problems.

\subsection{Omitted details of \texttt{UOB-REPS-EX}}
\label{appendix:UOB-REPS-EX}
\begin{algorithm2e}[!ht]
\begin{spacing}{0.9}
\label{UOB-REPS-EX}
\DontPrintSemicolon  
  \KwInitialize{state space $S$, action space $A$, episode number $T$, learning rate $\eta$, implicit exploration rate $\gamma$, explicit exploration rate $\beta$, confidence parameter $\delta$, Shannon Entropy $\{\Psi_h\}_{h\in [H]}$, and \texttt{Comp-UOB} as Algorithm 3 in \cite{jin2019learning}}
  \KwInitialize{epoch index $i=1$, confidence set $\mathcal{P}_1$ as the set of all transition functions, counters $N_0(s,a)=N_1(s,a)= M_0(s'|s,a)=M_1(s'|s,a)=0,\ \forall (s,a)$, occupancy measure $\hat{q}_1(s,a,s') = \frac{1}{|S_h|A||S_{h+1}|},\ \forall (s,a, s')$ and corresponding policy $\pi_1 = \pi^{\hat{q}_1}$}
  \KwInput{
   State exploration policies $\pi^s$, $\forall s$, (streaming) trajectories $\{s_h, a_h, \ell_t^+(s_h,a_h)\}_{h\in [H]}$ and clipping threshold $\{C_{t,h}\}_{h\in [H]}$ for $t\in [T]$
  }
  \KwOutput{
  (Streaming) policies $\pi_t$ for $t\in [T]$
  }
  \For{$t =1 \ \textbf{to} \ T$}
  {
    Send policy $\pi_t$ to \texttt{SCB-RL}. Receive trajectory $\{s_h, a_h, \ell_t^c(s_h,a_h)+C_{t,h}\}_{h\in [H]}$ and clipping threshold $\{C_{t,h}\}_{h\in [H]}$\;
    Compute upper occupancy bound:
    $u_t(s_h,a_h) = \texttt{Comp-UOB}(\pi_t, s_h, a_h, \mathcal{P}_{i}),\ \forall h\in [H]$\;
    Construct loss estimators:
    \begin{align*}
    \hat{\ell}_t(s,a) = \frac{\ell_t^c(s,a)+C_{t,h(s)}}{u_t(s,a)+\gamma}\mathbbm{1}\{s_{h(s)} = s, a_{h(s)} = a\},\ \forall (s,a)\in [S]\times [A]  
    \end{align*}
    Update counters: $N_i(s_h, a_h) \gets N_i(s_h, a_h) + 1,\ M_i(s_{h+1} | s_h, a_h) \gets M_i(s_{h+1} | s_h, a_h)  + 1,\ \forall h\in [H]$\;
    \If{$\exists h\in [H], \  N_i(s_h, a_h) \geq \max\{1, 2N_{i-1}(s_h,a_h)\}$}
    {
    Increase epoch index $i \gets i+1$\;
    Initialize new counters:  $N_i \gets N_{i-1},\ M_i\gets M_{i-1}$\;
    Update confidence set $\mathcal{P}_i$
    \begin{align*}
        \mathcal{P}_i = \bigg\{   \hat P : |\hat P(s'|s,a) - \bar P_i(s'|s,a) |\le \epsilon_i(s'|s,a),\\
        \forall (s,a,s')\in [S_h]\times [A]\times [S_{h+1}], h=0,\dots, H -1  \bigg\},
    \end{align*}
    where $\bar P_i(s'|s,a) = \frac{M_i(s'|s,a)}{\max\{ 1, N_i(s,a)   \}}$ and 
    \begin{align*}
        \epsilon_i(s'|s,a) = 4\sqrt{\frac{\bar P(s'|s,a) \ln\left(\frac{TSA}{\delta} \right)  }{\max\{1, N_i(s,a)-1 \}}} +\frac{28\ln\left(\frac{TSA}{\delta} \right) }{3\max\{1, N_i(s,a)-1 \}}
    \end{align*}
    }
    Update occupancy measure and policy, get $\widetilde{q}_{t+1}$ and set $\widetilde{\pi}_{t+1}\gets \pi^{\widetilde{q}_{t+1}}$
    \begin{align*}
        \widetilde{q}_{t+1} = \arg\min_{q\in\Delta(\mathcal{P}_i)} \left(\sum_{t} \langle q, \hat{\ell}_t \rangle + \sum_h  \frac{C_{t,h}}{\eta}\Psi_h(q)\right)\;
    \end{align*}\\
    Add extra exploration: $\pi_{t+1} = (1-\beta) \widetilde{\pi}_{t+1} +\beta \text{Uniform}(\pi^1,\dots, \pi^S)$\;
  }
\caption{\texttt{UOB-REPS-EX}: Upper Occupancy Bound Relative Entropy
Policy Search with Explicit Exploration}
\end{spacing}
\end{algorithm2e}

In this section, we introduce the algorithm \texttt{UOB-REPS-EX}, as illustrated in Algorithm~\ref{UOB-REPS-EX}. 
The algorithm is mainly the same to \texttt{UOB-REPS} in \cite{jin2019learning}, except for the following three differences.
First, \texttt{UOB-REPS-EX} uses the clipping loss with offset $\{\ell_t^c(s_h,a_h)+C_{t,h}\}_{h\in [H]}$ instead of $\{\ell_t(s_h,a_h)\}_{h\in [H]}$ as the input.
Secondly, in each episode, Algorithm~\ref{UOB-REPS-EX} applies FTRL with an adaptive learning rate to update the occupancy measure $\widetilde q_{t+1}$, instead of using OMD with a fixed learning rate.
Recall the definition of (negative) Shannon Entropy on an occupancy measure $q$ is 
\begin{align*}
    \Psi_h(q) = \sum_{s\in [S_h], a\in [A]} q(s,a)\ln \frac{1}{ q(s,a)},\ \forall h\in [H].
\end{align*}
Lastly, the policy output by \texttt{UOB-REPS-EX} is a mixture of its FTRL output policy $\widetilde \pi_{t+1}$ and the exploration policies from \texttt{RF-ELP}.
This step is to allow every state-action pair to have a probability of being visited, so that \texttt{SCB-RL} can perceive the change of loss scale and update the clipping threshold on time.

\subsection{Proof of Lemma~\ref{lem:rfelp}}
\label{appendix:lem:rfelp}
We first note that 
\begin{align*}
    \E\left[  \sum_h  r(s_h, a_h) |P, \pi \right] =  \E\left[ \sum_h \mathbbm{1}\{s_h=s\} |P, \pi \right]= q^{P,\pi}(s).
\end{align*}
Denoted by $q^s =  \max_{\pi\in \Pi} q^{P,\pi}(s)$, using the above, Lemma~\ref{MVPg} implies that
\begin{align*}
  q^s - q^{P,\pi^{s,N}}(s)\le \mathcal{\tilde O}\left( \sqrt{\frac{SA \textup{Var}^s }{N}}+\frac{SAH}{N}\right)\le \mathcal{\tilde O}\left( \sqrt{\frac{SA q^s }{N}} +\frac{SAH}{N}\right)
\end{align*}
holds with probability at least $1-\delta$, where the last inequality is due to $\textup{Var}^s\le q^s$.
By the appendix F.3.4 in \cite{zhang2023settling}, we can set $C = \mathcal{O}\left(\log^4(T)\log^2(SAH)\log(  \frac{ 1}{\delta})\right)$ such that $q^s - q^{P,\pi^{s,N}}(s)\le C\left( \sqrt{\frac{SA q^s }{N}} +\frac{SAH}{N} \right)$.
When $q^s> 9C^2\frac{SAH}{N}$,  we note that
\begin{align*}
     q^s>  2C\left( \sqrt{\frac{SA q^s }{N}} +\frac{SAH}{N} \right)
\end{align*}
and thus 
\begin{align*}
    q^{P,\pi^{s,N}}(s)\ge q^s-C\left( \sqrt{\frac{SA q^s }{N}} +\frac{SAH}{N} \right)\ge \frac{q^s}{2}.
\end{align*}
This completes the proof.

\subsection{Proof of Lemma~\ref{lemma:9}}
\label{appendix:lemma:9}

We start by stating two key technical lemmas from \cite{jin2019learning}. 
The first outlines the reliability of the confidence sets.
The second essentially describes how the confidence set shrinks over time.
\begin{lemma}(Lemma $2$ in \cite{jin2019learning})
\label{lem:key_lemma:1}
With probability at least $1-4\delta$, there is $P\in \mathcal{P}_i$ for all $i$.
\end{lemma}
\begin{lemma} (Lemma $4$ in \cite{jin2019learning})
\label{lem:key_lemma}
    With probability at least $1-\delta$, for any $h\in [H]$ and any collection of transition functions $\{P_t^s\}_{s\in [S]}$ such that $P_t^s\in \mathcal{P}_{i_t}$ for all $s\in [S]$, there is
    \begin{align*}
        \sum_{t=1}^T \sum_{s\in [S_h], a\in [A]} |q^{P_t^s, \pi_t}(s,a)-q^{P,\pi_t}(s,a)| \le \mathcal{O}\left( S\sqrt{AT\ln\left(\frac{TSA}{\delta}\right)}  \right)
    \end{align*} 
    \end{lemma}
    Recall $q_t = q^{P, \pi_t}$ and $\hat q_t = q^{\hat P_t, \pi_t}$.
Given the above lemma, we decompose the regret into 
\begin{align*}
    \sum_{t=1}^T \langle \ell_t^+, q_t-q^*\rangle = 
    \overbrace{\sum_{t=1}^T \langle \ell_t^+, q_t-\hat q_t\rangle}^{\textsc{Error}} + 
    \overbrace{\sum_{t=1}^T \langle \ell_t^+-\hat\ell_t, \hat q_t\rangle}^{\textsc{Bias}_1}
    + 
    \overbrace{\sum_{t=1}^T \langle \hat\ell_t, \hat q_t - q^\star\rangle}^{\textsc{Reg}}
    + 
    \overbrace{\sum_{t=1}^T \langle \hat\ell_t - \ell_t^+,  q^*\rangle}^{\textsc{Bias}_2}
\end{align*}

\paragraph{Bounding $\textsc{Error}$}:
By Lemma~\ref{lem:key_lemma}, we immediately obtain the following bound.
\begin{lemma}
\label{lem:bound_err}
With probability at least $1-\delta$, there is 
\begin{align*}
    \textsc{Error}\le  \mathcal{O}\left( \sum_{h\in [H]} \ell_{\infty, h} S\sqrt{AT\ln\left(\frac{TSA}{\delta}\right)} \right) 
\end{align*}
\end{lemma}

\paragraph{Bounding $\textsc{Bias}_1$}:
The high level idea of bounding $\textsc{Bias}_1$ is to show that $\hat\ell_t$ is not underestimating $\ell_t^+$ by too much, which is ensured due to the fact that the confidence set becomes more and more accurate for frequently visited state-action pairs.
\begin{lemma}
\label{lem:bound_bias_1}
    With probability at least $1-\delta$, there is 
    \begin{align*}
        \textsc{Bias}_1\le \mathcal{O}\left( \sum_{h\in [H]} \ell_{\infty, h} S\sqrt{AT\ln\left(\frac{SAT}{\delta}\right)} + \sum_{h\in [H]}  \gamma \ell_{\infty, h} S_h A T \right)
    \end{align*}
\end{lemma}

\paragraph{Bounding $\textsc{Reg}$}:
In this part, we build the proof based on the ideas of \cite{neu2015explore} and \cite{jin2019learning}.
The main challenge is that our loss estimator $\hat\ell_t$ corresponds to the policy $\pi_t$ rather than the FTRL output $\widetilde \pi_t$, which makes some regular proof tricks no longer applicable.
\begin{lemma}
    \label{lem:bound_reg}
    With probability at least $1-\delta$, there is
    \begin{align*}
            \textsc{Reg}\le \mathcal{O}\left( \frac{\ln(SA)}{\eta} \sum_{h\in [H]} \ell_{\infty,h} + \frac{\eta}{1-\beta} A T \sum_{h\in [H]} \ell_{\infty,h} S_h   + \frac{\ln\left(\frac{H}{\delta} \right)}{\gamma}  \sum_{h\in [H]} \ell_{\infty,h}+ \beta T \sum_{h\in [H]}  \ell_{\infty,h} \right).
    \end{align*}
\end{lemma}

\paragraph{Bounding $\textsc{Bias}_2$}:
$\textsc{Bias}_2$ can be bounded  via a direct application of Lemma~\ref{lem:IX}.
\begin{lemma}
\label{lem:bound_bias_2}
    With probability at least $1-\delta$, there is 
    \begin{align*}
        \textsc{Bias}_2\le \mathcal{O}\left( \frac{1}{\gamma}\sum_{h\in [H]}\ell_{\infty, h}\ln\left(\frac{H}{\delta} \right)   \right)
    \end{align*}
\end{lemma}

Summing up the above and setting $\eta =\gamma=\mathcal{O}\left(\sqrt{\frac{H\ln(SAT/\delta)}{SAT}}\right)$ and $\beta\ge 1/2$, we get
\begin{align*}
    \sum_{t=1}^T \langle \ell_t^+, q_t-q^*\rangle \le \mathcal{O}\left( \sum_{h\in [H]} \ell_{\infty, h} S\sqrt{AT\ln\left(\frac{SAT}{\delta}\right)} +\beta T  \sum_{h\in [H]} \ell_{\infty, h} \right)
\end{align*}
with probability $1-\delta$.

\subsection{Proof of Lemma~\ref{lemma:MDP_clip}}
\label{appendix:lemma:MDP_clip}
Recall $q^s = \max_{\pi\in \Pi}q^{P,\pi}(s) $ and $\ell_t' = |\ell_t-\ell_t^c|$.
By the clipping rule, there is $\ell'(s,a)\le |\ell_t(s,a)|\mathbbm{1}\{C_{t,h}< 2^i\}$ for every $(s,a)$ pair.
Fix $h\in [H]$, it suffices to prove 
\begin{align*}
        \sum_{s\in [S_h], a\in [A]} \sum_{t=1}^T |\ell_t(s,a)||q_t(s,a)-q^*(s,a)|
    \mathbbm{1}\left\{ q^s> \mathcal{\tilde O}(SAH/\xi T)  \right\}\mathbbm{1}\left\{C_{t,h}< 2^i\right\}&\\
    \le \mathcal{\tilde O}\left(\frac{ \ell_{\infty,h} SA}{\beta}  \right)&
\end{align*}
Define $K:=\arg\min_{j\in \mathbb{N}}\left\{  \ell_{\infty, h} \le 2^j \right\}$.
Define $\ell_t^i(s,a) = \ell_t(s,a)\mathbbm{1}\{2^{i-1}< |\ell_t(s,a)|\le 2^{i}\}\mathbbm{1}\{ q^s> \mathcal{\tilde O}(SAH/\xi T)  \} $.
We note that
\begin{align*}
     &\sum_{s\in [S_h], a\in [A]} \sum_{t=1}^T |\ell_t(s,a)||q_t(s,a)-q^*(s,a)|
    \mathbbm{1}\left\{ q^s> \mathcal{\tilde O}(SAH/\xi T)  \right\}\mathbbm{1}\left\{C_{t,h}< 2^i\right\}\\
    \le& \sum_{i=-\infty}^K\sum_{s\in [S_h], a\in [A]}\sum_{t=1}^T |\ell_t^i(s,a)||q_t(s,a)-q^*(s,a)|\mathbbm{1}\left\{C_{t,h}< 2^i\right\}\\
    \le&  \sum_{i=-\infty}^K\sum_{s\in [S_h], a\in [A]}\sum_{t=1}^T |\ell_t^i(s,a)|q_s \mathbbm{1}\left\{C_{t,h}< 2^i\right\} \\
    \le& \sum_{i=-\infty}^K 2^i \sum_{t=1}^T \mathbbm{1}\left\{C_{t,h}< 2^i\right\} \sum_{s\in [S_h], a\in [A]} q_s\mathbbm{1}\left\{\ell_t^i(s,a)\not = 0\right\} .
\end{align*}
For brevity, we denote by $X_t^i = \sum_{s\in [S_h], a\in [A]} q_s\mathbbm{1}\{\ell_t^i(s,a)\not = 0\}$.
By Lemma~\ref{lem:rfelp}, for any $s\in [S_h]$, if $q_s\mathbbm{1}\left\{\ell_t^i(s,a)\not = 0\right\} \not=0$, state $s$ can be well explored by \texttt{RF-ELP} with high probability.
As shown in Algorithm~\ref{UOB-REPS-EX}, the exploration policy has probability at least $\mathcal{\tilde O}(\beta /SA)$ to be played in episode $t$ for every state.
Thus, the algorithm is able to observe the outlier and update $C_{t+1,h}$ to $2^i$ with probability at least $\mathcal{\tilde O}(\beta X_t^i/SA) $.
Thus, for every integer $m\ge 1$, we have
\begin{align*}
    \P\left\{ \sum_{t=1}^T \mathbbm{1}\left\{C_{t,h}< 2^i\right\} X_t^i \ge \sum_{t=1}^m X_t^i \right\}&\le \prod_{t=1}^m \left( 1- \mathcal{\tilde O}(\beta X_t^i/SA)  \right)\\
    &\le \mathcal{\tilde O}\left( \left( 1- \frac{\beta}{SA}  \right)^{\sum_{t=1}^m X_t^i} \right).
\end{align*}
This implies that with probability at least $1-\delta$, there is
\begin{align*}
    \sum_{t=1}^T \mathbbm{1}\left\{C_{t,h}< 2^i\right\} \sum_{s\in [S_h], a\in [A]} q_s\mathbbm{1}\left\{\ell_t^i(s,a)\not = 0\right\}\le \mathcal{\tilde O}\left(\frac{SA}{\beta}\right ).
\end{align*}
Then, using the same idea of Lemma~\ref{High_prop:2}, we can achieve a high probability union bound on all $i\le K$, i.e., with probability at least $1-\delta$
\begin{align*}
    &\sum_{i=-\infty}^K 2^i \sum_{t=1}^T \mathbbm{1}\left\{C_{t,h}< 2^i\right\} \sum_{s\in [S_h], a\in [A]} q_s\mathbbm{1}\left\{\ell_t^i(s,a)\not = 0\right\} \\
    \le& \sum_{i=-\infty}^K 2^i \mathcal{\tilde O}\left(\frac{SA}{\beta}\right )\le 2^{K+1} \mathcal{\tilde O}\left(\frac{SA}{\beta}\right ) = \mathcal{\tilde O}\left(\frac{\ell_{\infty, h}SA}{\beta}\right ),
\end{align*}
which completes the proof.

\subsection{Proof of Lemma~\ref{lem:bound_err}}
\begin{align*}
    \sum_{t=1}^T \langle \ell_t^+, q_t-\hat q_t\rangle &\le \sum_{h\in [H]}\left( 
\max_{s\in [S_h], a\in [A]}\ell_t^+(s,a) \left( \sum_{t=1}^T \sum_{s\in [S_h], a\in [A]} |q^{\hat P_t, \pi_t}(s,a)-q_t(s,a)| \right) \right) \\
&\le \sum_{h\in [H]}\left( 
\max_{s\in [S_h], a\in [A]}\ell_t^+(s,a) \left( \sum_{t=1}^T \sum_{s\in [S_h], a\in [A]} |q^{ P_t^s, \pi_t}(s,a)-q_t(s,a)| \right) \right) \\
&\le \mathcal{O}\left( \sum_{h\in [H]} C_{T,h} S\sqrt{AT\ln\left(\frac{TSA}{\delta}\right)} \right) \\
&\le \mathcal{O}\left( \sum_{h\in [H]} \ell_{\infty, h} S\sqrt{AT\ln\left(\frac{TSA}{\delta}\right)} \right) 
\end{align*}
The second inequality is by setting $\hat P_t = P_t^s\in \mathcal{P}_{i_t}$ as in Lemma~\ref{lem:key_lemma}.
The third and last inequalities are due to $\ell_t^+(s, a)\le 2C_{t,h(s)}\le 2C_{T,h(s)}$ and $C_{T,h}\le 2\max_{s\in [S_h], a\in [A]}\ell_t^{+}(s,a) = 4\ell_{\infty, h}$ for all $h\in [H]$.

\subsection{Proof of Lemma~\ref{lem:bound_bias_1}}
We first note that 
\begin{align*}
    \E\left[\hat\ell_t(s,a) \right] = \frac{\ell_t^c(s,a)+C_{t,h(s)}}{u_t(s,a)+\gamma}q_t(s,a)  = \frac{q_t(s,a)}{u_t(s,a)+\gamma}\ell_t^+(s,a).
\end{align*}
Then
\begin{align*}
    \sum_{t=1}^T \langle \ell_t^+-\hat\ell_t, \hat q_t\rangle &= \sum_{t=1}^T \sum_{h\in [H]}\sum_{s\in [S_h], a\in [A]} \hat q_t(s, a) \left(\ell_t^+(s,a) -  \hat\ell_t(s,a)  \right)\\
    & = \sum_{t=1}^T \sum_{h\in [H]}\sum_{s\in [S_h], a\in [A]} \hat q_t(s, a) \left(\ell_t^+(s,a) -  \E\left[\hat\ell_t(s,a) \right] \right) \\
    &+ \sum_{t=1}^T \sum_{h\in [H]}\sum_{s\in [S_h], a\in [A]} \hat q_t(s, a) \left( \E\left[\hat\ell_t(s,a) \right] -  \hat\ell_t(s,a)  \right)
\end{align*}
For the first term, there is 
\begin{align*}
    &\sum_{t=1}^T \sum_{h\in [H]}\sum_{s\in [S_h], a\in [A]} \hat q_t(s, a) \left(\ell_t^+(s,a) -  \E\left[\hat\ell_t(s,a) \right] \right) 
    \\
    =& \sum_{t=1}^T \sum_{h\in [H]}\sum_{s\in [S_h], a\in [A]} \hat q_t(s, a) \ell_t^+(s,a) \left(1 -  \frac{q_t(s,a)}{u_t(s,a)+\gamma} \right)\\
    \le & 3 \sum_{h\in [H]} \ell_{\infty, h} \sum_{t=1}^T \sum_{s\in [S_h], a\in [A]}  \frac{\hat q_t(s,a)}{u_t(s,a)+\gamma} \left(u_t(s,a)+\gamma-q_t(s,a) \right)\\
    \le& 3 \sum_{h\in [H]} \ell_{\infty, h} \sum_{t=1}^T \sum_{s\in [S_h], a\in [A]}   |u_t(s,a)-q_t(s,a)|+ 3\sum_{h\in [H]}  \gamma \ell_{\infty, h} S_h A T.
\end{align*}
Recall 
\begin{align*}
       u_t(s,a) = \pi_t(a|s)\max_{\hat P\in \mathcal{P}_{i_t}} q^{\hat P, \pi_t}(s),
\end{align*}
the last inequality is due to $\hat q_t(s,a)\le u_t(s,a)$.
Moreover, since $ q^{P_t^x, \pi_t}(s,a)=\pi_t(a|s)q^{P_t^x, \pi_t}(s)\le u_t(s,a) $ for all $(s,a)\in [S]\times [A]$, it suffices to bound $ \sum_{t=1}^T \sum_{s\in [S_h], a\in [A]}   |u_t(s,a)-q_t(s,a)|$ by Lemma~\ref{lem:key_lemma}.
Thus we can conclude
\begin{align*}
    &\sum_{t=1}^T \sum_{h\in [H]}\sum_{s\in [S_h], a\in [A]} \hat q_t(s, a) \left(\ell_t^+(s,a) -  \E\left[\hat\ell_t(s,a) \right] \right) 
    \\
    \le& \mathcal{O}\left( \sum_{h\in [H]} \ell_{\infty, h} S\sqrt{AT\ln\left(\frac{SAT}{\delta}\right)} + \sum_{h\in [H]}  \gamma \ell_{\infty, h} S_h A T \right).
\end{align*}

For the second term, notice that $|\sum_{h\in [H]}\sum_{s\in [S_h], a\in [A]} \hat q_t(s, a)  \hat\ell_t(s,a) | \le 3\sum_{h\in [H]} \ell_{\infty,h} $ for all $t\in [T]$.
Using Azuma's inequality, with probability at least $1-\delta$, we have 
\begin{align*}
    \sum_{t=1}^T \sum_{h\in [H]}\sum_{s\in [S_h], a\in [A]} \hat q_t(s, a) \left( \E\left[\hat\ell_t(s,a) \right] -  \hat\ell_t(s,a)  \right)\le \mathcal{O}\left(  \sum_{h\in [H]} \ell_{\infty,h} \sqrt{T\ln \frac{1}{\delta}} \right).
\end{align*}
Summing up the two terms and resize $\delta$ completes the proof.

\subsection{Proof of Lemma~\ref{lem:bound_reg}}
We start by decomposing \textsc{Reg} into
\begin{align*}
    \textsc{Reg} &= (1-\beta) \sum_{t=1}^T \langle \hat\ell_t, \widetilde q_t - q^\star\rangle + \beta \sum_{t=1}^T \langle \hat\ell_t, q^{\hat P_t, \text{Uniform}(\pi^1,\dots,\pi^S)} - q^\star\rangle\\
    &\le (1-\beta) \sum_{t=1}^T \langle \hat\ell_t, \widetilde q_t - q^\star\rangle +  \beta \sum_{t=1}^T \langle \hat\ell_t - \ell_t^+, q^{\hat P_t, \text{Uniform}(\pi^1,\dots,\pi^S)} \rangle + \beta \sum_{t=1}^T \langle  \ell_t^+, q^{\hat P_t, \text{Uniform}(\pi^1,\dots,\pi^S)} \rangle\\
    &\le  (1-\beta) \sum_{t=1}^T \langle \hat\ell_t, \widetilde q_t - q^\star\rangle +  \beta \sum_{t=1}^T \langle \hat\ell_t - \ell_t^+, q^{\hat P_t, \text{Uniform}(\pi^1,\dots,\pi^S)} \rangle + 3 \beta T\sum_{h\in [H]} \ell_{\infty,h} 
\end{align*}
To bound the first and second term, we propose a variant of Lemma 11 in \cite{jin2019learning}.
\begin{lemma}
    \label{lem:IX}
    For any sequence of functions $\alpha_{1}, \dots, \alpha_T$ such that $\alpha_t \in [0, 2\gamma]^{S\times A}$ and $\mathcal{F}_t$-measurable for all $t\in [T]$, with probability at least $1 -\delta$, there is 
    \begin{align*}
       \sum_{t=1}^T\sum_{s\in [S],a\in [A]} \alpha_t(s,a) \left(\hat \ell_t(s,a) - \frac{q_t(s,a)}{u_t(s,a)}\ell_t^+(s,a)\right) \le  \mathcal{O}\left( \sum_{h\in [H]} \ell_{\infty, h} \ln\left(\frac{H}{\delta}\right) \right).
    \end{align*}
\end{lemma}
Without loss of generality, in the following, we assume that $u_t(s,a)\ge q_t(s,a)$ for all $(s,a)\in [S]\times [A]$, which holds true with probability at least $1-\delta$.
Using Lemma~\ref{lem:IX}, since $q^{\hat P_t, \text{Uniform}(\pi^1,\dots,\pi^S)}$ is independent to $\ell_t$ and thus be $\mathcal{F}_t$-measurable, we can immediately bound the second term by 
\begin{align*}
    \beta \sum_{t=1}^T \langle \hat\ell_t - \ell_t^+, q^{\hat P_t, \text{Uniform}(\pi^1,\dots,\pi^S)} \rangle &= \frac{\beta}{2\gamma}\sum_{t=1}^T \langle \hat\ell_t - \ell_t^+,2\gamma q^{\hat P_t, \text{Uniform}(\pi^1,\dots,\pi^S)} \rangle\\
    &\le \frac{\beta}{2\gamma} \sum_{h\in [H]}\ell_{\infty,h}\ln\left(\frac{H}{\delta}\right)
\end{align*}
with probability at least $1-\delta$.
It suffices to focus on the first term.
By standard analysis of FTRL algorithm, there is 
\begin{align*}
    \sum_{t=1}^T \langle \hat\ell_t, \widetilde q_t - q^\star\rangle \le \sum_{h\in [H]} \frac{C_{T,h}}{\eta}\ln(SA) + \sum_{t=1}^T \sum_{h\in [H]} \frac{\eta}{C_{t,h}} \sum_{s\in [S_h], a\in [A]} \widetilde q_t(s,a) \hat\ell_t^2(s,a).
\end{align*}
We further note that
\begin{align*}
   \widetilde q_t(s,a) \hat\ell_t^2(s,a) \le \widetilde q_t(s,a)\frac{\ell_t^+(s,a)}{u_t(s,a)+\gamma} \hat\ell_t(s,a)\le 2C_{t,h} \frac{\widetilde q_t(s,a)}{u_t(s,a)+\gamma} \hat\ell_t(s,a).
\end{align*}
Different to the proof of \cite{jin2019learning}, we cannot immediately conclude $\widetilde q_t(s,a)/(u_t(s,a)+\gamma)\le 1$.
This is because $u_t(s,a)$ is the upper bound of $q^{\hat P_t, \pi_t}(s,a)$ rather than $q^{\hat P_t, \widetilde \pi_t}(s,a)$.
Here we prove by showing
\begin{align*}
    u_t(s,a)\ge q^{\hat P_t, \pi_t}(s,a) \ge (1-\beta)q^{\hat P_t, \widetilde \pi_t}(s,a),
\end{align*}
which implies that $\widetilde q_t(s,a)/(u_t(s,a)+\gamma)\le 1/(1-\beta)$.
Thus we have
\begin{align*}
    \sum_{t=1}^T \langle \hat\ell_t, \widetilde q_t - q^\star\rangle \le \sum_{h\in [H]} \frac{C_{T,h}}{\eta}\ln(SA) + \frac{2\eta}{1-\beta} \sum_{t=1}^T \sum_{h\in [H]}  \sum_{s\in [S_h], a\in [A]}  \hat\ell_t(s,a).
\end{align*}
Applying Lemma~\ref{lem:IX} obtains
\begin{align*}
    &\sum_{t=1}^T \langle \hat\ell_t, \widetilde q_t - q^\star\rangle\\
    \le & \sum_{h\in [H]} \frac{C_{T,h}}{\eta}\ln(SA) + \frac{2\eta}{1-\beta} \sum_{t=1}^T \sum_{h\in [H]}  \sum_{s\in [S_h], a\in [A]} \frac{q_t(s,a)}{u_t(s,a)} \ell_t(s,a) + \sum_{h\in [H]} \frac{\ell_{\infty,h}}{2\gamma}\ln\left(\frac{H}{\delta}\right)\\
    \le & \sum_{h\in [H]} \frac{2\ell_{\infty,h}}{\eta}\ln(SA) + \frac{2\eta}{1-\beta} \sum_{h\in [H]} \ell_{\infty,h} S_h A T  + \sum_{h\in [H]} \frac{\ell_{\infty,h}}{2\gamma}\ln\left(\frac{H}{\delta}\right) 
\end{align*}
with probability at least $1-\delta$.
Combining with the above and resize $\delta$ we finally get
\begin{align*}
    \textsc{Reg}\le \mathcal{O}\left( \frac{\ln(SA)}{\eta} \sum_{h\in [H]} \ell_{\infty,h} + \frac{\eta}{1-\beta} A T \sum_{h\in [H]} \ell_{\infty,h} S_h   + \frac{\ln\left(\frac{H}{\delta} \right)}{\gamma}  \sum_{h\in [H]} \ell_{\infty,h}+ \beta T \sum_{h\in [H]}  \ell_{\infty,h} \right)
\end{align*}
completing the proof.

\subsection{Proof of Lemma~\ref{lem:bound_bias_2}}
Assuming $u_t(s,a)\ge q_t(s,a)$ for all $(s,a)\in [S]\times [A]$ be true.
Using Lemma~\ref{lem:IX}, we have
\begin{align*}
    \sum_{t=1}^T \langle \hat\ell_t - \ell_t^+,  q^*\rangle&\le \sum_{t=1}^T \sum_{s\in [S], a\in [A]} q^*(s,a)\left(\frac{q_t(s,a)}{u_t(s,a)}\ell_t^+(s,a) - \ell_t^+(s,a) \right)  + \frac{1}{2\gamma}\sum_{h\in [H]}\ell_{\infty, h}\ln\left(\frac{H}{\delta} \right)\\
    &\le \frac{1}{2\gamma}\sum_{h\in [H]}\ell_{\infty, h}\ln\left(\frac{H}{\delta} \right)
\end{align*}
with probability at least $1-\delta$.
Resize $\delta$ completes the proof.

\subsection{Proof of Lemma~\ref{lem:IX}}
We note that
\begin{align*}
    &\sum_{t=1}^T\sum_{s\in [S],a\in [A]} \alpha_t(s,a) \left(\hat \ell_t(s,a) - \frac{q_t(s,a)}{u_t(s,a)}\ell_t^+(s,a)\right) \\
    &= \sum_{h\in [H]} 3\ell_{\infty, h}  \sum_{t=1}^T \sum_{s\in [S_h],a\in [A]} \alpha_t(s,a) \left( \frac{\hat \ell_t(s,a)}{3\ell_{\infty, h}} - \frac{q_t(s,a)}{u_t(s,a)}\frac{\ell_t^+(s,a)}{3\ell_{\infty, h}}\right).
\end{align*}
Now $\ell_t^+(s,a)/3\ell_{\infty, h}$ is within $[0,1]$ for all $t\in [T]$ and $(s,a)\in [S]\times [A]$.
By the results of Lemma 11 in \cite{jin2019learning}, there is
\begin{align*}
    \sum_{t=1}^T \sum_{s\in [S_h],a\in [A]} \alpha_t(s,a) \left( \frac{\hat \ell_t(s,a)}{3\ell_{\infty, h}} - \frac{q_t(s,a)}{u_t(s,a)}\frac{\ell_t^+(s,a)}{3\ell_{\infty, h}}\right)\le \ln\left(\frac{H}{\delta} \right)
\end{align*}
for all $h\in [H]$ with probability at least $1-\delta$.
Thus we have
\begin{align*}
    \sum_{t=1}^T\sum_{s\in [S],a\in [A]} \alpha_t(s,a) \left(\hat \ell_t(s,a) - \frac{q_t(s,a)}{u_t(s,a)}\ell_t^+(s,a)\right)\le  \mathcal{O}\left( \sum_{h\in [H]} \ell_{\infty, h} \ln\left(\frac{H}{\delta}\right) \right),
\end{align*}
which completes the proof.

\subsection{Omitted details of Remark~\ref{remark:3}}
\label{appendix:remark:3}
In this section, we describe how to reduce the regret of \texttt{SCB-RL} to $\tilde{\mathcal{O}}(\sum_{h\in [H]}\ell_{\infty, h} S\sqrt{AT})$.
Recall \texttt{RF-ELP}, we fix the number of episodes used to find an exploration policy for state $s$ as \(\mathcal{O}(\sqrt{SAT})\). 
This is actually not necessary, that is, if the exploration algorithm has already found a good exploration policy for state $s$, it should stop searching and take the policy as output. 
In this case, the number of episodes used to find an exploration policy will be independent of \(T\). 
Inspired by this, we design \texttt{RF-ELP-ES}, as illustrated in Algorithm~\ref{RF-ELP-ES}.
We will elucidate the details of the algorithm in the following section.

By Lemma~\ref{MVPg} and Lemma~F.3.4 in \cite{zhang2023settling}, there exists $C = \mathcal{O}(\log^3(T)\log^2(SAH) \\  \log(  \frac{ 1}{\delta}))$ such that 
\begin{align*}
    q^s - q^{P,\pi^{s, N}}(s)\le C\left( \sqrt{\frac{SA q^s }{N}}+ \frac{SAH}{N}  \right)
\end{align*}
for all $N\ge 1$ with probability at least $1-\delta$. 
Taking the above as a quadratic function of $\sqrt{q^s}$, we have
\begin{align*}
    \sqrt{q^s}\le \sqrt{q^{P,\pi^{s, N}}(s)} + 2C\sqrt{\frac{SAH}{N}},
\end{align*}
which immediately implies
\begin{align}
\label{extension:1}
    q^s - q^{P,\pi^{s, N}}(s)&\le 3C^2\left( \sqrt{\frac{SA  q^{P,\pi^{s, N}}(s) }{N}}+ \frac{SAH}{N}  \right).
\end{align}
Given $q^{P,\pi^{s, N}}(s) = \E[\sum_{t=1}^N \mathbbm{1}_t\{s |P, \pi_t^s\}]/N$,
by empirical Bernstein's inequality, there exists $C' = \mathcal{O}(\log(  \frac{ T}{\delta}))$ such that with probability at least $1-\delta$, for all $N\ge 1$,  there is
\begin{align}
\label{extension:2}
    \left| q^{P,\pi^{s, N}}(s) - \frac{\sum_{t=1}^N \mathbbm{1}_t\{s |P, \pi_t^s\}}{N}  \right|\le C'\left( 
 \frac{\sqrt{\sum_{t=1}^N \mathbbm{1}_t\{s |P, \pi_t^s\}}}{N}+ \frac{1}{N}
 \right).
\end{align}
Combining inequalities~(\ref{extension:1}) and (\ref{extension:2}), it suffices to show that
\begin{align}
\label{extension:3}
    q^s\le \frac{\sum_{t=1}^N \mathbbm{1}_t\{s |P, \pi_t^s\}}{N}  + C{''}\left(  \frac{\sqrt{SA\sum_{t=1}^N \mathbbm{1}_t\{s |P, \pi_t^s\}}}{N}+ \frac{SAH}{N}  \right)
\end{align}
where $C{''} = \mathcal{O}(\log^7(T)\log^4(SAH)\log^3(  \frac{ 1}{\delta}))$.
Furthermore, by inequality~\ref{extension:2}, we further have 
\begin{align*}
     q^{P,\pi^{s, N}}(s)  \ge \frac{\sum_{t=1}^N \mathbbm{1}_t\{s |P, \pi_t^s\}}{N}- C'\left( 
 \frac{\sqrt{\sum_{t=1}^N \mathbbm{1}_t\{s |P, \pi_t^s\}}}{N}+ \frac{1}{N}
 \right).
\end{align*}
This means 
\begin{align*}
    \frac{q^s}{q^{P,\pi^{s, N}}(s)}\le \frac{
{\sum_{t=1}^N \mathbbm{1}_t\{s |P, \pi_t^s\}}  + C{''}\left(  {\sqrt{SA\sum_{t=1}^N \mathbbm{1}_t\{s |P, \pi_t^s\}}}+ {SAH}  \right)    
    }{
    {\sum_{t=1}^N \mathbbm{1}_t\{s |P, \pi_t^s\}}- C'\left( 
 {\sqrt{\sum_{t=1}^N \mathbbm{1}_t\{s |P, \pi_t^s\}}}+ 1
 \right)
    }.
\end{align*}
When $\sum_{t=1}^N \mathbbm{1}_t\{s |P, \pi_t^s\}\ge 9{C^{''}}^2 SAH$, it suffices to say that ${q^s}/{q^{P,\pi^{s, N}}(s)}\le 4$ with probability at least $1-2\delta$, that is, policy $\pi^{s, N}$ is good enough to explore state $s$, thus we can stop \texttt{RF-ELP-ES} in advance.

The last question is how large does $N$ need to make $\sum_{t=1}^N \mathbbm{1}_t\{s |P, \pi_t^s\}\ge 9{C^{''}}^2 SAH$ with probability at least $1-\delta$.
Taking inequality~(\ref{extension:3}) as a quadratic function of $\sqrt{\sum_{t=1}^N \mathbbm{1}_t\{s |P, \pi_t^s\}}$, we note that
\begin{align*}
    \sqrt{\sum_{t=1}^N \mathbbm{1}_t\{s |P, \pi_t^s\}}\ge \frac{-C^{''}\sqrt{SA} +\sqrt{{C^{''}}^2 SA +4Nq^s-4SAH  }}{2}.
\end{align*}
\texttt{RF-ELP-ES} will end when 
\begin{align*}
    \frac{-C^{''}\sqrt{SA} +\sqrt{{C^{''}}^2 SA +4Nq^s-4SAH  }}{2}\ge 3{C^{''}}^2 \sqrt{SAH}.
\end{align*}
The above holds if $N\ge 16 {C^{''}}^2 {SAH}/{q_s}$.
Therefore, \texttt{RF-ELP-ES} requires at most $\tilde{\mathcal{O}}({SAH}/{q_s})$ episodes to find an exploration policy for every state $s$.

\begin{algorithm2e}[!t]
\label{RF-ELP-ES}
\DontPrintSemicolon  
  \KwInput{State $s$; Upper bound of exploration episodes number $N$ }
  \KwOutput{Policy $\pi\in \Pi$; Exploration episodes number $N'$}
  Initialize reward: $r^s(s', a') \leftarrow \mathbbm{1}\{s'=s\}$ for all $(s',a')\in [S]\times A$\;
  Run \texttt{MVP} \citep{zhang2023settling} $N'$ episodes, where $N' = \min\{N, \arg\min_{M} \sum_{t=1}^M \mathbbm{1}_t\{s |P, \pi_t^s\}\ge 9{C^{''}}^2 SAH\}$, get policies $\{\pi^{s}_{1},\dots,\pi^s_{N'}\} \gets \texttt{MVP}(r^s, N')$, set $\pi^{s, N'}(\cdot|s) \leftarrow \text{Uniform}(\pi^{s}_{1},\dots,\pi^s_{N'})$\;
  Set $\pi^{s, {N'}}(\cdot|s) \leftarrow \text{Uniform}(A)$\;
  Return $\pi^{s, {N'}}$, $N'$\;
\caption{Reward free exploration in RL with Early Stopping (\texttt{RF-ELP-ES})}
\end{algorithm2e}

\paragraph{Regret analysis}
Denote by 
\begin{align*}
    q_{\min} = \min_{s\in [S]}\left\{ q^s   \right\}  = \min_{s\in [S]}\left\{\max_{\pi\in \Pi} q^{P,\pi}(s)   \right\}.
\end{align*}
Consider $T\ge \tilde{\mathcal{O}}\left(\frac{SAH^2}{q_{\min}^2}\right)$ such that $\xi T\ge 16 {C^{''}}^2 {SAH}/{q_s}$ for all $s\in [S]$.
In this case, every state is thoroughly explored, thus the term $\tilde{\mathcal{O}}\left(\frac{ \sum_{h\in [H]}\ell_{\infty,h}  S_h S H A}{\xi}\right)$ in the regret can be eliminated.
Moreover, we can reduce the error incurred by the exploration phase from $\mathcal{O}\left( \sum_{h\in [H]}\ell_{\infty, h} \xi S T \right)$ to $\tilde{\mathcal{O}}\left( \sum_{h\in [H]}\ell_{\infty, h} \frac{S^2AH}{q_{\min}} \right)$ since $\texttt{RF-ELP-ES}$ operates at most $\tilde{\mathcal{O}}\left( \frac{S^2AH}{q_{\min}} \right)$ episodes.
Combining with other terms in the regret, we finally have
\begin{align*}.
\mathcal{R}(T)\le \mathcal{\tilde O}\left(\sum_{h\in [H]} \ell_{\infty, h}   \left[  S\sqrt{AT} +\beta T + \frac{   SA}{\beta} +\frac{S^2AH}{q_{\min}} \right]  \right).
\end{align*}
Setting $\beta = \mathcal{O}(\sqrt{SA/T})$ concludes the proof.
\end{document}